\documentclass[twocolumn]{IEEEtran} 
\usepackage[T1]{fontenc}
\usepackage{color}
\usepackage{mathrsfs}
\usepackage{amsmath}
\usepackage{amsthm} 
\usepackage{amssymb}
\usepackage{graphicx}
\usepackage{cite}
\usepackage{makecell}
\usepackage{algorithm}
\usepackage{algorithmic}
\usepackage{fancyhdr}
\usepackage{diagbox}
\usepackage{bm}
\usepackage[unicode=true,
 bookmarks=true,bookmarksnumbered=true,bookmarksopen=true,bookmarksopenlevel=1,
 breaklinks=false,pdfborder={0 0 0},pdfborderstyle={},backref=false,colorlinks=false]
 {hyperref}
\hypersetup{
 pdftitle={Your Title},
 pdfauthor={Your Name},
 pdfpagelayout=OneColumn, pdfnewwindow=true, pdfstartview=XYZ, plainpages=false}

\usepackage{multirow} 
\usepackage{xcolor}

\allowdisplaybreaks[4]

\usepackage[caption=false,font=footnotesize]{subfig}

\IEEEoverridecommandlockouts

\usepackage{geometry}
\begin{document}
\bibliographystyle{IEEEtran}

\title{\textcolor{black}{Hierarchical Task Offloading and Trajectory Optimization in Low-Altitude Intelligent Networks Via Auction and Diffusion-based MARL}

\author{\IEEEauthorblockN{Jiahao You, Ziye Jia, \textit{Member, IEEE,} Can Cui, Chao Dong, \IEEEmembership{Senior Member,~IEEE,} \\
Qihui Wu, \IEEEmembership{Fellow,~IEEE,} and Zhu Han, \IEEEmembership{Fellow,~IEEE}}

\thanks{Jiahao You, Can Cui, Chao Dong, and Qihui Wu are with the Key Laboratory of Dynamic Cognitive System of Electromagnetic Spectrum Space, Nanjing University of Aeronautics and Astronautics, Nanjing 211106, China (e-mail: yjiahao@nuaa.edu.cn, cuican020619@nuaa.edu.cn, dch@nuaa.edu.cn, wuqihui@nuaa.edu.cn).

Ziye Jia is with the Key Laboratory of Dynamic Cognitive System of Electromagnetic Spectrum Space, Nanjing University of Aeronautics and Astronautics, Nanjing 211106, China, and also with the National Mobile Communications Research Laboratory, Southeast University, Nanjing, Jiangsu, 211111, China (e-mail: jiaziye@nuaa.edu.cn).

Zhu Han is with the University of Houston, Houston, TX 77004 USA, and also with the Department of Computer Science and Engineering, Kyung Hee University, Seoul 446-701, South Korea (e-mail: hanzhu22@gmail.com).}}}

\maketitle

\begin{abstract}
The low-altitude intelligent networks (LAINs) emerge as a promising architecture for delivering low-latency and energy-efficient edge intelligence in dynamic and infrastructure-limited environments. By integrating unmanned aerial vehicles (UAVs), aerial base stations, and terrestrial base stations, LAINs can support mission-critical applications such as disaster response, environmental monitoring, and real-time sensing. However, these systems face key challenges, including energy-constrained UAVs, stochastic task arrivals, and heterogeneous computing resources.
To address these issues, we propose an integrated air-ground collaborative network and formulate a time-dependent integer nonlinear programming problem that jointly optimizes UAV trajectory planning and task offloading decisions. The problem is challenging to solve due to temporal coupling among decision variables.
Therefore, we design a hierarchical learning framework with two timescales. At the large timescale, a Vickrey-Clarke-Groves auction mechanism enables the energy-aware and incentive-compatible trajectory assignment. At the small timescale, we propose the diffusion-heterogeneous-agent proximal policy optimization, a generative multi-agent reinforcement learning algorithm that embeds latent diffusion models into actor networks. Each UAV samples actions from a Gaussian prior and refines them via observation-conditioned denoising, enhancing adaptability and policy diversity.
Extensive simulations show that our framework outperforms baselines in energy efficiency, task success rate, and convergence performance.
\end{abstract}

\section{Introduction} \label{s1}
In recent years, the deployment of low-altitude intelligent networks (LAINs) has emerged as a promising solution to deliver low-latency and energy-efficient edge intelligence in dynamic and infrastructure-limited environments \cite{EETO_2022,3DTO_2022,SFCD_2025}. LAINs integrate unmanned aerial vehicles (UAVs), aerial base stations (ABSs), and terrestrial base stations (TBSs) into a cooperative edge computing framework that supports mission-critical applications such as disaster relief, environmental monitoring, aerial surveillance, and real-time sensing~\cite{UTOT_2022,AETO_2024, Sustainable_Wang_2024}. Among these components, UAVs play a central role due to their high mobility, rapid deployment capability, and the ability to establish line-of-sight communication links, making them well suited for supplementing or replacing ground infrastructure in remote or disrupted environments \cite{NFV_cyl_2025,Joint_zhao_2025,Chen_2024}.

However, the practical realization of LAINs introduces several critical challenges arising from the inherent limitations of UAVs and the dynamics of edge environments. UAVs typically operate with limited onboard energy and computational capacity~\cite{EATD_2022, G2S_2025_arxiv, Disrupting_2025_arxiv}, which restricts their flight duration and ability to handle complex workloads independently. In addition, edge environments are highly dynamic, characterized by unpredictable task arrivals, heterogeneous service demands, and uneven spatial distribution of resources. These factors jointly necessitate continuous adaptation and coordination strategies. Coordinating UAV swarms under such conditions requires intelligent and scalable algorithms that can make real-time decisions while jointly considering spatial-temporal dependencies, energy budgets, and computational constraints~\cite{CWDU_2023,TVBM_2024,DROA_2025,Cooperative_2024}.

To overcome these limitations, we propose the integrated air-ground collaborative network (IAGCN), in which UAVs, ABSs, and TBSs jointly provide mobile edge computing services. In this framework, UAVs travel to designated task areas and process tasks locally or offload them to nearby cooperative nodes, including other UAVs, the ABS, or ground TBSs. This collaborative architecture enhances the flexibility of resource utilization and improves load balancing across heterogeneous edge nodes.
In the IAGCN, the ABS operates at a high altitude to ensure wide-area communication coverage and resilient coordination. Compared to TBSs, which provide stable and high-throughput computing services in fixed locations, the ABS supports dynamic task offloading when nearby TBSs are unavailable or overloaded. Furthermore, the ABS facilitates system-wide coordination and assists in maintaining connectivity across task areas through its superior communication reach.
The primary goal is to maximize the number of successfully completed tasks while minimizing the overall system energy consumption. However, the optimization process is challenging due to the intricate coupling between UAV-task assignment decisions and control actions such as trajectory adjustment and offloading scheduling. Although all decision variables are binary, the problem exhibits strong temporal dependencies and is formulated as a time-dependent integer nonlinear programming (INLP) problem, which is difficult to solve by conventional methods in dynamic multi-agent environments.

To address these challenges, we design a hierarchical learning framework that operates on two distinct timescales. At the large timescale, we perform the trajectory optimization through a Vickrey-Clarke-Groves (VCG) auction mechanism \cite{VCG_2005,VCG_2023, Two_based_2024}. The VCG mechanism ensures truthful participation and allocates UAVs in an energy-efficient and globally optimal manner based on task deadlines, spatial constraints, and UAV status. At the small timescale, we develop diffusion-heterogeneous-agent proximal policy optimization (D-HAPPO), a multi-agent reinforcement learning (MARL) algorithm that incorporates latent diffusion models (LDMs) \cite{LDM_2022, Aerial_2025_sun, Generative_2025_Wang} into actor networks. By treating action selection as a conditional generative process, LDMs enable each UAV to sample actions from a Gaussian latent prior and iteratively refine them through a time-aware denoising network conditioned on observation features. This generative modeling enhances policy expressiveness, improves robustness under uncertainties, and enables more diverse decision-making strategies. Our main contributions are summarized as follows:
\begin{itemize}
    \item We propose the IAGCN that leverages the spatial agility of UAVs and the computational capabilities of ABSs and TBSs to deliver resilient edge intelligence in dynamic low-altitude scenarios.
    \item We formulate a joint optimization problem for trajectory planning and task offloading, subject to key system constraints such as task deadlines, UAV mobility, and energy budgets.
    \item We design a two-timescale hierarchical solution. At the large timescale, a VCG auction-based assignment algorithm allocates UAVs to task areas. At the small timescale, we introduce D-HAPPO, a generative MARL algorithm that embeds LDMs into actor networks to enhance the policy generalization and adaptability.
    \item We conduct extensive simulations to evaluate the proposed framework. Results demonstrate that our method achieves superior energy efficiency, task success rate, and convergence performance compared to the MARL baselines.
\end{itemize}

The rest of this paper is organized as follows. Section~\ref{s2} reviews related works on UAV networking, energy optimization, and auction/diffusion-based learning methods. Section~\ref{s3} presents the system architecture and problem formulation. Section~\ref{s4} details the proposed VCG auction algorithm and D-HAPPO method. Section~\ref{s5} discusses simulation results and performance comparisons. Section~\ref{s6} concludes the paper and outlines future directions.

\section{Related Work}\label{s2}
This section reviews three critical research areas related to UAV-enabled LAINs: architecture and applications, energy optimization, and auction and diffusion models.

\subsection{Architecture and Applications of LAINs}
LAINs emphasize the deployment of UAVs in near-ground airspace to support high-density operations and improve resource efficiency. UAVs are increasingly adopted across domains such as logistics, disaster management, and environmental monitoring, and prompt growing interests in designing robust, cost-effective, and high-throughput architectures tailored to low-altitude environments~\cite{Computational_Cao_2024,Revolutionizing_Mahboob_2024,Cooperative_2025_low,Airspace_2024}.
For example,~\cite{Sensitive_2024} investigates multi-UAV cooperations in data collection for wireless sensor networks, aiming to optimize the age of information in densely deployed scenarios. Similarly,~\cite{Multi_2024} proposes a cooperative model for UAV-assisted computing that improves load balancing and reduces latency in infrastructure-limited regions.
In emergency settings,~\cite{Secure_2023} introduces a blockchain-based UAV-fog architecture for secure data sharing and efficient computation. To address GPS signal limitations,~\cite{Navigation_2024} surveys advanced multi-sensor fusion techniques that enhance UAV navigation accuracy.
In the context of distributed intelligence,~\cite{MultiAgent_2024} applies deep reinforcement learning to optimize multi-UAV trajectory coordination for edge computing. In urban vehicular networks,~\cite{VEC_2024} incorporates UAVs into vehicular edge computing systems to alleviate roadside unit congestion. Moreover,~\cite{ECE_2024} presents an elastic collaborative intelligence framework that enables UAVs to cooperatively execute deep learning inference under network instability.

\subsection{Energy Optimization in LAINs}

Due to the limited onboard battery capacity, energy efficiency remains a major constraint in UAV operations. Various approaches aim to improve the endurance and optimize resource allocation while maintaining performance.
For instance,~\cite{TOEP_2024} proposes a joint optimization framework that combines deep reinforcement learning and convex approximation for task offloading and trajectory control in hybrid UAV-edge systems. \cite{ELEJ_2024} jointly optimizes computation, transmission, and trajectory decisions via an alternating iteration scheme.
To balance fairness and efficiency,~\cite{APAF_2023} develops a parameterized deep Q-network that coordinates UAV offloading and flight operations. For sustainability,~\cite{EETO_2024} explores UAV trajectory planning with wireless energy transfer from high-altitude platforms. A multi-objective approach is introduced in~\cite{MODO_2024} to jointly optimize coverage utility and energy usage.
Other studies explore novel architectures. For example,~\cite{JECT_2024} proposes the synchronized task scheduling in UAV-enabled transportation systems, while~\cite{RISA_2024} incorporates reconfigurable intelligent surfaces to enhance energy harvesting. \cite{TSSF_2024} focuses on optimizing UAV-based wireless power transfer efficiency and energy consumption.

\subsection{Auction and Diffusion Models in LAINs}
The auction and diffusion mechanisms provide promising solutions for resource allocation and distributed learning in UAV-assisted networks. Auction-based methods support efficient and incentive-compatible task distribution, while diffusion-based models enable scalable decentralized coordination.
For example,~\cite{CFBG_2023} proposes a coalition auction approach where sensor nodes bid for UAV services, minimizing information delay and ensuring social welfare. In~\cite{OAAT_2024}, a two-stage auction framework addresses task allocation and UAV scheduling to improve efficiency in service disruption scenarios.
Diffusion models are also applied to distributed UAV systems.~\cite{Diffusion_du_2024} proposes a decentralized diffusion process where UAVs exchange local estimates to achieve global consensus under dynamic topologies. The privacy-preserving auction mechanisms, such as Ptero~\cite{IMOT_2024}, combine auctions with task offloading while maintaining UAV privacy and energy efficiency.
Hybrid frameworks are explored in~\cite{IOW_2023}, which uses a Stackelberg-auction game for marine UAV coordination. In~\cite{DNN_2025}, multi-agent diffusion reinforcement learning enhances task offloading in vehicular edge environments. Additionally,~\cite{MOAC_2025} utilizes generative diffusion models for secure and energy-efficient UAV communication.

Although the prior studies provide insights into UAV networking and task optimization, most adopt single-timescale frameworks that fail to jointly address long-term planning and real-time control. The integration of auction mechanisms with learning-based policies is rarely considered, and the use of diffusion models for generative decision-making in heterogeneous, hybrid action spaces remains largely unexplored.

\begin{table*}[t]
\caption{Key Notations}
\label{T1}
\begin{center}
\begin{tabular}{|m{4cm}<{\centering}|m{11cm}|}
\hline
\textbf{Symbol} & \textbf{Description} \\
\hline
$\mathbb{U}, \mathbb{B}, \mathbb{H}, \mathbb{T}$ & Set of UAVs, TBSs, the ABS, and time slots, respectively. \\
\hline
$u, b, h, t$ & Indices for UAVs, TBSs, ABS, and time slots, respectively. \\
\hline
$l_u(t), l_b(t), l_h(t)$ & Location of UAV $u$, TBS $b$, and ABS $h$ at time slot $t$, respectively. \\
\hline
$\mathcal{D}_m$ & Task area $m$ centered at $(x_m, y_m)$ with radius $r_m$. \\
\hline
$a_{m,n}$ & Task $n$ in area $m$, characterized by size $d_{m,n}$, workload $c_{m,n}$, and deadline $t_{m,n}^{dl}$. \\
\hline
$\mathcal{Q}_{m}^{ta}(t), \mathcal{Q}_u^{uav}(t), \mathcal{Q}_\alpha(t)$ & Task queues at task area $m$, UAV $u$, and node $\alpha \in \{\mathbb{U}, \mathbb{B}, \mathbb{H}\}$, respectively. \\
\hline
$x_{m,n}^u(t), x_{m,n}^{u,\alpha}(t)$ & Binary variables for local processing or offloading of task $a_{m,n}$ by UAV $u$ at time slot $t$. \\
\hline
$z_{u,m}(t), z^{r}_{u,m}(t)$ & Binary variables indicating UAV $u$'s connection and proximity to task area $\mathcal{D}_m$ at time slot $t$. \\
\hline
$R_{u,\alpha}^{u2\alpha}(t), B_{u,\alpha}(t)$ & Transmission rate and allocated bandwidth between UAV $u$ and node $\alpha$. \\
\hline
$f_u, f_\alpha$ & CPU frequency of UAV $u$ and node $\alpha$, respectively. \\
\hline
$E_{m,n}^u(t), E_{m,n}^{u,\alpha}(t)$ & Energy consumption for UAV $u$ to process task $a_{m,n}$ locally or offload to node $\alpha$. \\
\hline
$E_u^{fly}(t), E_u(t)$ & Flight energy and total energy consumption of UAV $u$ at time slot $t$. \\
\hline
$E^{s}, \eta$ & System-wide total energy consumption and energy efficiency metric. \\
\hline
$t_{m,n}^{cp}, t_{m,n}^{cp,\alpha}(t)$ & Completion time for task $a_{m,n}$ under local execution or offloading to node $\alpha$. \\
\hline
$\mathcal{S}_{m}(t)$ & Set of successfully completed tasks in area $\mathcal{D}_m$ at time slot $t$. \\
\hline
$\bm{x}, \bm{z}$ & Joint decision variables for offloading and UAV-task area association. \\
\hline
$U_u(m), b_u(m)$ & Utility and bid of UAV $u$ for task area $m$ in the auction process. \\
\hline
$\text{Price}_u$ & VCG price paid by UAV $u$ reflecting its marginal system utility. \\
\hline
$\mathbf{z}_t, \mathbf{z}_0, \epsilon$ & Noisy latent variable, initial Gaussian prior, and random noise in the diffusion model. \\
\hline
$\mathcal{L}_{\text{diff}}, D_\theta$ & Diffusion actor network loss and denoising network. \\
\hline
$M^{i_{1:m}}(s,\bm{a})$ & Compound advantage function used in D-HAPPO training. \\
\hline
\end{tabular}\label{table:notation}
\end{center}
\end{table*}

\section{System Model}\label{s3}
We consider the IAGCN, which consists of an ABS, multiple UAVs in the air, and several TBSs. The key notations used in this paper are summarized in Table~\ref{table:notation}.

\subsection{Network Model}
\begin{figure}[!t]
\centering
\includegraphics[width=7cm]{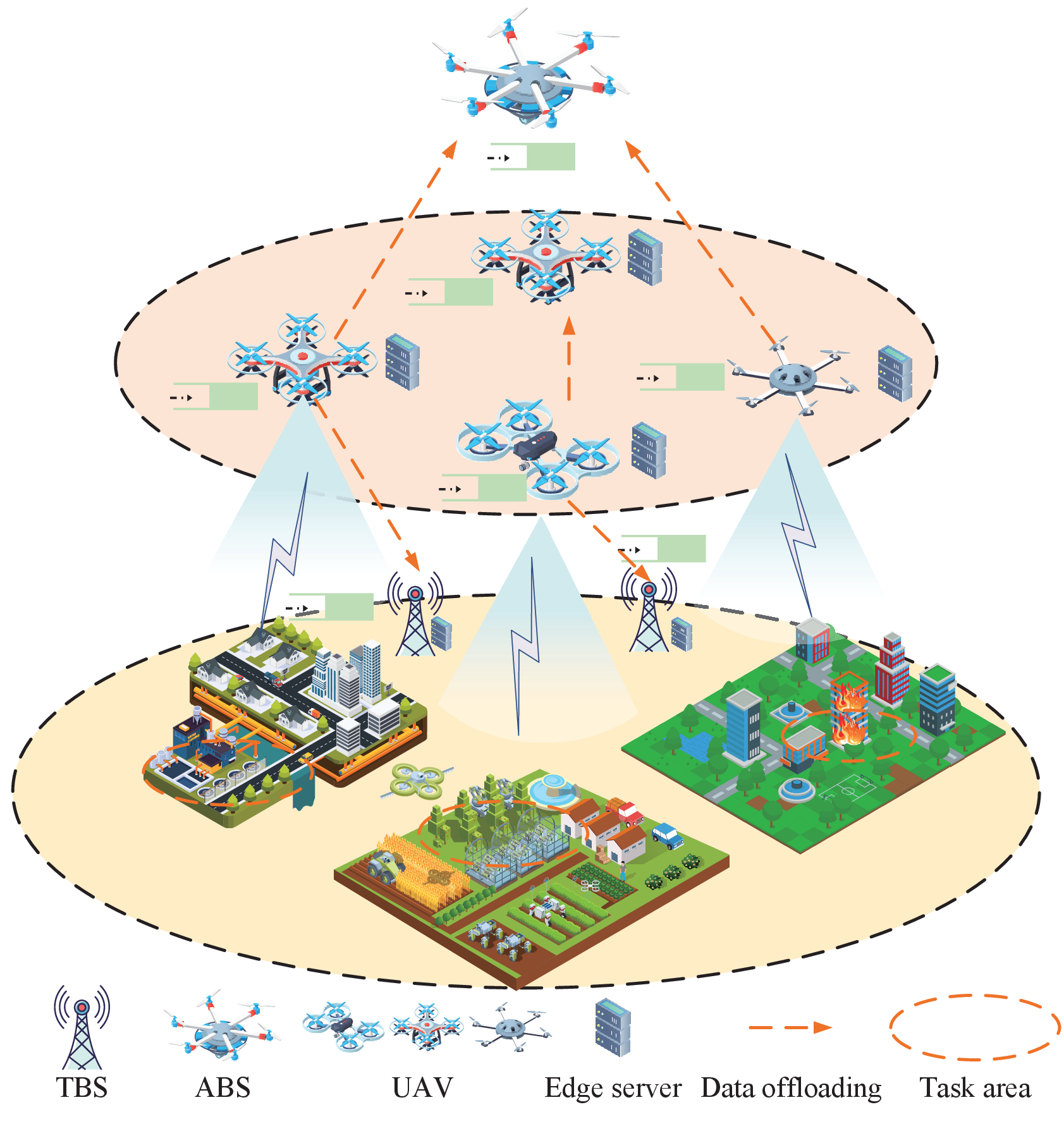}
\caption{System Model Overview.} 
\label{f1}
\end{figure}

As illustrated in Fig. \ref{f1}, the IAGCN is designed to enable UAVs to operate within designated task areas to perform complex missions such as power line inspection, agricultural monitoring, and emergency rescue operations. The ABS serves as a high-altitude node, providing wide coverage and robust computational capabilities. TBSs, equipped with significant processing power, act as stable resource hubs for both computation and communication.
Tasks are generated within assigned task areas when UAVs operate in these regions. Due to the limited computational capabilities and energy resources of UAVs, these tasks are often offloaded to nearby TBSs or the ABS for collaborative processing. Furthermore, UAVs can redistribute tasks to nearby UAVs in a single-hop manner, enabling dynamic and efficient resource sharing.

The set of UAVs is denoted as $\mathbb{U} = \{1, 2, \dots, U\}$, the set of TBSs is denoted as $\mathbb{B} = \{1, 2, \dots, B\}$, and the ABS is represented as $\mathbb{H} = \{H\}$. The operational time is divided into discrete time slots, denoted as $\mathbb{T} = \{1, 2, \dots, T\}$, each with a duration of $\tau$.
The position of UAV $u$ at time slot $t$ is expressed as $l_u(t) = (x_u(t), y_u(t), z_u(t))$, where $x_u(t)$, $y_u(t)$, and $z_u(t)$ represent its coordinates in the horizontal, vertical, and altitude dimensions, respectively. Similarly, the position of TBS $b$ is denoted as $l_b(t) = (x_b(t), y_b(t), z_b(t))$, and the position of the ABS is given as $l_h(t) = (x_h(t), y_h(t), z_h(t))$.
    
\begin{figure}[!t]
\centering
\includegraphics[width=7cm]{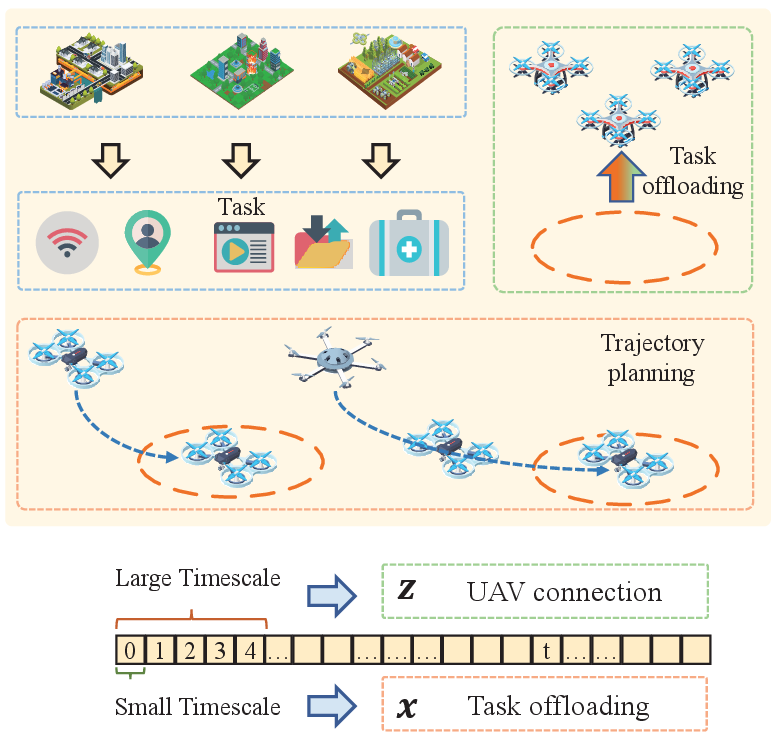}
\caption{Two timescale optimization.} 
\label{f2}
\end{figure}
    
\subsection{Time-Scale Separation}
To manage the complexity of jointly optimizing UAV trajectory planning and task offloading decisions, we adopt a time-scale separation strategy. This hierarchical approach decomposes the original problem into two interrelated subproblems, enabling efficient handling of large-scale spatial coordination and responsive offloading decisions.

\begin{itemize} 
    \item \textbf{Large timescale}: UAV movements are constrained by physical limits on speed and acceleration, and are typically planned over extended intervals or planning epochs. 
    \item \textbf{Small timescale}: Task offloading decisions are made dynamically at each time slot, based on instantaneous observations such as channel conditions, queue states, and computational resource availability. 
\end{itemize}

\subsubsection{Large timescale}
At the large timescale, we model UAV trajectories to minimize the flight energy while ensuring adequate task coverage. This requires establishing models for task area definitions, UAV mobility patterns, and associated energy consumption over long-term horizons.

\paragraph{Task Area Model}
In the proposed system, task areas are modeled as circular regions on a two-dimensional plane. Each task area $\mathcal{D}_m$ is centered at position $l_m(t) = (x_m, y_m)$ and defined by its radius $r_m$ as:
\begin{equation}
\mathcal{D}_m = \{(x, y) \mid (x - x_m)^2 + (y - y_m)^2 \leq r_m^2 \}.
\end{equation}

Each area maintains a task queue $\mathcal{Q}_{m}^{\text{ta}}(t)$ that is initialized at time slot $t=0$ with $N_m$ tasks:
\begin{equation}
\mathcal{Q}_{m}^{\text{ta}}(0) = \{a_{m,1}, a_{m,2}, \dots, a_{m,N_m}\}.
\end{equation}

Each task $a_{m,n}$ is characterized by a tuple $\{d_{m,n}, c_{m,n}, t_{m,n}^{\text{dl}}\}$, where $d_{m,n}$ denotes the data size, $c_{m,n}$ is the required computation in CPU cycles per bit, and $t_{m,n}^{\text{dl}}$ is the task deadline in time slots:
\begin{equation}
a_{m,n} = \{d_{m,n}, c_{m,n}, t_{m,n}^{\text{dl}}\}.
\end{equation}

When multiple tasks are assigned to a UAV, they are processed in the order of arrival.

\paragraph{UAV Mobility Model}
To enhance the energy efficiency, UAVs are assumed to maintain a fixed cruising altitude $H$, thereby avoiding frequent ascents and descents. As such, we model the UAV movement as horizontal planar motion over designated task areas \cite{Joint_Optimization_Zhao_2025, Tradeoff_Between_Zhan_2024,UAV_Enabled_Xu_2018 }. According to the energy dissipation model in~\cite{al_Ahmed_2023}, the flight energy consumption of UAV $u$ during time slot $t$ comprises two components: the \emph{hovering} power consumption $p_u^{\mathrm{hov}}$ and the \emph{moving} power consumption $p_u^{\mathrm{mov}}(v_u(t))$. The hovering power is given by:
\begin{equation}
p_u^{\mathrm{hov}} = \sqrt{\frac{(M_u g)^3}{2 \pi \rho_u p_u \vartheta}},
\end{equation}
where $M_u$ is the UAV mass, $g$ is the gravitational acceleration, $\rho_u$ is the propeller radius, $p_u$ is the number of propellers, and $\vartheta$ denotes the air density.

When UAV $u$ moves at speed $v_u(t)$, additional motion-induced power is incurred~\cite{Energy_Mu_2021}:
\begin{equation}
p_u^{\mathrm{mov}}(v_u(t)) = \frac{v_u(t)}{v_u^{\mathrm{max}}} \left(p_u^{\mathrm{max}} - p_u^{\mathrm{stop}}\right),
\end{equation}
where $v_u^{\mathrm{max}}$ is the maximum speed, $p_u^{\mathrm{max}}$ is the corresponding power consumption at full speed, and $p_u^{\mathrm{stop}}$ is the idle power consumption while hovering ($v_u(t)=0$).

Let $\delta_u(t)$ denote the horizontal distance traveled by UAV $u$ in time slot $t$, calculated as $\delta_u(t)=\|l_u(t+1) - l_u(t)\|$. Then, the total flight energy consumption is expressed as:
\begin{equation}
E_u^{\mathrm{fly}}(t) = \left(p_u^{\mathrm{hov}} + p_u^{\mathrm{mov}}(v_u(t))\right) \frac{\delta_u(t)}{v_u(t)}.
\end{equation}

To capture the UAV-task area association, we define a binary decision variable \( z_{u,m}(t) \in \{0,1\}\), which indicates whether UAV \( u \) is connected to task area \( \mathcal{D}_m \) at time slot \( t \):
\begin{equation}
z_{u,m}(t) = 
\begin{cases}
1, & \text{if UAV } u \text{ is connected to task area } \mathcal{D}_m, \\
0, & \text{otherwise}.
\end{cases}
\label{e7}
\end{equation}

\subsubsection{Small Timescale}

Let \( \kappa_{\alpha} \) denote the energy consumption coefficient for executing CPU/GPU cycles at node \( \alpha \), where \( \alpha \in \{\mathbb{U}, \mathbb{B}, \mathbb{H}\} \) represents another UAV, a TBS, or the ABS. 

To determine whether UAV \( u \) physically reaches task area \( \mathcal{D}_m \) at time slot \( t \), we define an availability indicator variable \( z^r_{u,m}(t) \in \{0,1\} \), given by:
\begin{equation}
z^r_{u,m}(t) =
\begin{cases}
1, & \text{if } (x_u(t) - x_m)^2 + (y_u(t) - y_m)^2 \leq r_m^2, \\
0, & \text{otherwise}.
\end{cases}
\label{zr_def}
\end{equation}

The communication between UAVs, TBSs, and the ABS is essential for task offloading. The following communication models govern the data transfer rates for offloading tasks.
\paragraph{UAV Communication Model}
In the IAGCN, the topology remains quasi-static within each time slot, meaning that the channel state information (CSI) is approximately constant. The CSI is updated at the beginning of each time slot based on the latest parameters. The communication models between UAVs, TBSs, and the ABS determine the achievable transmission rates, which impact the task offloading completion time.

\begin{equation}
\begin{split}
\xi_{u,h}(t) &= \frac{\zeta_{L}-\zeta_{NL}}{1+a \exp \{-b [\varsigma_{u,h}(t) -a] \}} \\
& + 20 \lg\left(\frac{4\pi f_c\|l_u(t)-l_h(t)\| }{C}\right) + \zeta_{NL},
\end{split}
\end{equation}
where $\varsigma_{u,h}(t)$ is the elevation angle, $f_c$ is the carrier frequency, and $C$ is the speed of light. 
The achievable transmission rate from UAV $u$ to the ABS is:
\begin{equation} R_{u,h}^{u2a}(t) = B_{u,h}^{u2a}(t) \log_2 \left( 1 + \frac{P_u(t) \xi_{u,h}(t)}{N_G} \right), \end{equation}
where $B_{u,h}^{u2a}(t)$ is the allocated bandwidth, $P_u(t)$ is the UAV's transmit power, and $N_G$ is the noise power.
The path loss between UAV $u$ and node $\sigma$ (which can be either TBS $b$ or UAV $u'$) is modeled as:

\begin{equation}
G_{u,\sigma}^{u2\sigma}(t) = G_0 (l_u(t) - l_{\sigma}(t))^{-2},
\end{equation}
where $\sigma \in \{\mathbb{B}, \mathbb{U}\}$ represents the destination node (TBS or another UAV).
The achievable transmission rate from UAV $u$ to node $\sigma$ is 
\begin{equation}
R_{u,\sigma}^{u2\sigma}(t) = B_{u,\sigma}^{u2\sigma}(t) \log_2 \left( 1 + \frac{P_u(t) G_{u,\sigma}^{u2\sigma}(t)}{N_G} \right),
\end{equation}
in which $B_{u,\sigma}^{u2\sigma}(t)$ represents the available bandwidth, $P_u(t)$ is the UAV's transmit power, and $N_G$ is the noise power.

\paragraph{Local Processing}
The processing or offloading status of each task \(a_{m,n}\) at time slot \(t\) is described using the following binary variable:
\begin{equation}\label{e13}
x_{m,n}^u(t) = 
\begin{cases} 
1, & \text{if } a_{m,n} \text{ is processed locally by UAV } u, \\
0, & \text{otherwise}.
\end{cases}
\end{equation}

For task \(a_{m,n}\) processed locally on UAV \(u\), the completion time equals the sum of scheduling overhead, the UAV's queueing delay, and the service time, i.e.,
\begin{equation}
t_{m,n}^{\text{cp},u} = t_{m,n}^{\text{start}} + t_{m,n}^{\text{wait},u} + \frac{d_{m,n} c_{m,n}}{f_u},
\end{equation}
where \( t_{m,n}^{\text{start}} \) denotes the time slot at which task \( a_{m,n} \) is scheduled for execution. 
The waiting time \(t_{m,n}^{\text{wait},u}\) equals the sum of the remaining service times of the tasks already in UAV \(u\)'s queue, i.e.,
\begin{equation}
t_{m,n}^{\text{wait},u} = \sum_{a_{i,j} \in \mathcal{Q}_u^{\text{uav}}(t)} \frac{d_{i,j} c_{i,j}}{f_u},
\end{equation}
where \( \mathcal{Q}_u^{\text{uav}}(t) \) represents the local task queue of UAV \( u \) at time slot \( t \), and \( f_u \) is the CPU frequency of UAV \( u \), measured in cycles per second. 
Each UAV \( u \in \mathbb{U} \) maintains a local task queue \( \mathcal{Q}_u^{\text{uav}}(t) \) at time slot \( t \), which stores tasks assigned to or processed by UAV \( u \). The queue is updated as follows:
\begin{equation}
\begin{aligned} 
\mathcal{Q}_u^{\text{uav}}(t+1) = &\Big(\mathcal{Q}_u^{\text{uav}}(t) \cup \{a_{m,n} \mid x_{m,n}^u(t) = 1\}\Big) \\
&\setminus \Big(\{a_{m,n} \mid t_{m,n}^{\text{cp},u} \leq t_{m,n}^{\text{dl}} \text{ or } t_{m,n}^{\text{cp},u} > t_{m,n}^{\text{dl}}\}\Big),
\end{aligned}
\end{equation}
where \( x_{m,n}^u(t) = 1 \) indicates that task \( a_{m,n} \) is processed locally by UAV \( u \) at time slot \( t \).
The energy consumption \cite{UEE_2024,MLTE_2025} for processing task \( a_{m,n} \) locally on UAV \( u \) is given by :
\begin{equation}
E_{m,n}^u(t) = \kappa_u^{\text{uav}} (f_u)^2 d_{m,n}c_{m,n},
\end{equation}
where \( \kappa_u^{\text{uav}} \) is the effective switched capacitance coefficient of the CPU in UAV \( u \).

\paragraph{Offloading to Another Node}
To represent the offloading to other nodes, we define a unified offloading decision variable for UAVs, TBSs, and the ABS. Let \(\alpha\) represent the destination node, where \(\alpha \in \{\mathbb{U}, \mathbb{B}, \mathbb{H}\}\), and the offloading decision for task \(a_{m,n}\) at time slot \(t\) is:
\begin{equation}\label{e18}
x_{m,n}^{u,\alpha}(t) = 
\begin{cases} 
1, & \text{if task } a_{m,n} \text{ is offloaded } \text{ to node } \alpha, \\
0, & \text{otherwise}.
\end{cases}
\end{equation}

Offloading completion time is the sum of transmission time, queueing delay at the destination node, and remote service time.
If the task \( a_{m,n} \) is offloaded from UAV \( u \) to another node \( \alpha \), the completion time is 
\begin{equation}
t_{m,n}^{\text{cp},\alpha} = t_{m,n}^{\text{start}} + \frac{d_{m,n}}{R^{u2\alpha}_{u,\alpha}(t)} + t_{m,n}^{\text{wait},\alpha}(t) + \frac{d_{m,n} c_{m,n}}{f_{\alpha}},
\end{equation}
where the waiting time \( t_{m,n}^{\text{wait},\alpha}(t) \) is calculated as:
\begin{equation}
t_{m,n}^{\text{wait},\alpha}(t) = \sum_{a_{i,j} \in \mathcal{Q}_{\alpha}(t)} \frac{d_{i,j} c_{i,j}}{f_{\alpha}},
\end{equation}
where \( f_\alpha \) denotes the CPU frequency (in cycles per second) of node \( \alpha \). The completion time \( t_{m,n}^{\text{cp}} \) for a task \( a_{m,n} \) can be expressed as:
\begin{equation}
t_{m,n}^{\text{cp}} = x_{m,n}^u(t) t_{m,n}^{\text{cp},u}(t) + \sum_{\alpha \in \{\mathbb{U}, \mathbb{B}, \mathbb{H}\}} x_{m,n}^{u,\alpha}(t) t_{m,n}^{\text{cp},\alpha}(t),
\end{equation}
where \( x_{m,n}^{u,\alpha}(t) \) is a binary variable indicating whether task \( a_{m,n} \) is offloaded from UAV \( u \) to node \( \alpha \) at time slot \( t \).
When a task is offloaded from UAV \( u \) to another node \( \alpha \), the task queues are updated as 
\begin{equation}
\begin{aligned}
\mathcal{Q}_u^{\text{uav}}(t+1) = &\mathcal{Q}_u^{\text{uav}}(t) \setminus \{a_{m,n} \mid x_{m,n}^{u,\alpha}(t) = 1\},
\end{aligned}
\end{equation}
and 
\begin{equation}
\begin{aligned}
\mathcal{Q}_{\alpha}(t+1) = &\Big(\mathcal{Q}_{\alpha}(t) \cup \{a_{m,n} \mid x_{m,n}^{u,\alpha}(t) = 1\}\Big) \\
&\setminus \Big(\{a_{m,n} \mid t_{m,n}^{\text{cp},\alpha} \leq t_{m,n}^{\text{dl}} \text{ or } t_{m,n}^{\text{cp},\alpha} > t_{m,n}^{\text{dl}}\}\Big),
\end{aligned}
\end{equation}

The energy consumption for offloading task \( a_{m,n} \) from UAV \( u \) to another node \( \alpha \) is:
\begin{equation}
E_{m,n}^{u,\alpha}(t) = P_u(t) \frac{d_{m,n}}{R_{u,\alpha}^{u2\alpha}(t)
} + \kappa_{\alpha}(f_{\alpha})^2 d_{m,n}c_{m,n},
\end{equation}
where \( \kappa_{\alpha} \) is the effective switched capacitance coefficient of node \( \alpha \).
To ensure that each task is processed either locally or offloaded to a single destination, the following constraint is introduced:
\begin{equation}
\begin{aligned}
\sum_{t \in \mathbb{T}} &\bigg( x_{m,n}^u(t) + \sum_{\alpha \in \{\mathbb{U}, \mathbb{B}, \mathbb{H}\}} x_{m,n}^{u,\alpha}(t) \bigg) \leq 1. \label{e25}
\end{aligned}
\end{equation}

In our model, only tasks completed before their deadlines are considered successful and are included in the task success set \( \mathcal{S}_m(t) \). Tasks missing their deadlines are neither retried nor rescheduled and are implicitly treated as dropped, in line with real-time system constraints.
To formalize this, we define the task success set \( \mathcal{S}_m(t) \), which contains all tasks \( a_{m,n} \) in task area \( \mathcal{D}_m \) that complete execution before their respective deadlines:
\begin{equation}
\mathcal{S}_{m}(t) = \{a_{m,n} \mid t_{m,n}^{\text{cp}} \leq t_{m,n}^{\text{dl}}\}.
\end{equation}

\subsubsection{Total Energy Consumption}
The total energy consumption of UAV $u \in \mathbb{U}$ during a time slot $t$ consists of three main components: flight energy consumption, computation energy consumption, and communication energy consumption.

The processing energy consumption for a task \( a_{m,n} \) in time slot \( t \) is given by:
\begin{equation}
\begin{aligned}
E^{\text{proc}}_u(t) = &\ x_{m,n}^u(t) E_{m,n}^u(t) + \sum_{\alpha \in \{\mathbb{U}, \mathbb{B}, \mathbb{H}\}} x_{m,n}^{u,\alpha}(t) E_{m,n}^{u,\alpha}(t).
\end{aligned}
\end{equation}

The total energy consumption of UAV \( u \) during time slot \( t \) is the sum of its flight energy consumption and processing energy consumption:
\begin{equation}
E_u(t) = E^{\text{fly}}_u(t) + E^{\text{proc}}_u(t).
\end{equation}

The system-wide energy consumption considers the cumulative energy consumption of all UAVs in the network over the entire operational period:
\begin{equation}
E^{\text{s}} = \sum_{u \in \mathbb{U}} \sum_{t \in \mathbb{T}} E_u(t).
\end{equation}

To evaluate system efficiency, we define the energy efficiency metric \( \eta \) as the amount of successfully processed task data per unit of energy consumed, measured in bits per joule. This metric reflects the system's ability to maintain high task throughput under energy constraints while balancing productivity and sustainability. A higher bits-per-joule value indicates more efficient use of limited onboard energy.
\begin{equation}
\eta = \frac{ \sum_{m \in \mathcal{D}} \sum_{a_{m,n} \in \mathcal{S}_{m}} d_{m,n} }{ E^{\text{s}} },
\end{equation}
where \( \mathcal{D} \) denotes the set of all task areas, \( \mathcal{S}_{m} \) represents the set of tasks in area \( \mathcal{D}_m \) that are successfully completed before their deadlines, and \( E^{\text{s}} \) is the total energy consumption of all UAVs during the entire operation.

\subsection{Problem Formulation}
The goal of the optimization problem is to maximize the system's energy efficiency and the proportion of tasks completed within their deadlines across the IAGCN. Specifically, we jointly optimize the UAV-to-task-area connections \(\bm{z}\) and the task offloading decisions \(\bm{x}\), under various constraints.

\begin{subequations}
\begin{align}
\mathscr{P}0: \quad
& \max_{\bm{z}, \bm{x}} 
  \eta + \beta \sum_{m \in \mathcal{D}} 
     \frac{\sum_{t \in \mathbb{T}} \left|\mathcal{S}_{m}(t)\right|}{N_m}
  \notag \\[4pt]
& \text{s.t.} 
  \quad (\ref{e7}), (\ref{e13}), (\ref{e18}), (\ref{e25}), \notag \\[2pt]
& z^{r}_{u,m}(t) \leq z_{u,m}(t),  \forall u \in \mathbb{U}, m \in \mathcal{D}, t \in \mathbb{T}, \label{Za} \\[2pt]
& x_{m,n}^u(t) \leq z^{r}_{u,m}(t), \forall u \in \mathbb{U}, m, n \in \mathcal{D}, t \in \mathbb{T}, \label{Zb} \\[2pt]
& x_{m,n}^{u,\alpha}(t) \leq z^{r}_{u,m}(t), \quad 
  \forall u \in \mathbb{U},\, \alpha \in \{\mathbb{U}, \mathbb{B}, \mathbb{H}\}, \notag \\ 
& \qquad\qquad\qquad\qquad\qquad m, n \in \mathcal{D},\, t \in \mathbb{T}, \label{Zc}
\end{align}
\end{subequations}
where \(\eta\) denotes the system-wide energy efficiency, defined as the total successfully processed task data (in bits) divided by the total energy consumption. \(\left|\mathcal{S}_{m}(t)\right|\) represents the number of tasks completed on time in task area \(\mathcal{D}_m\) at time slot \(t\), and \(N_m\) is the total number of tasks in that area. The summation term thus captures the average task success rate across time and space, and \(\beta\) is a weighted factor that balances task reliability with energy efficiency.
Constraint~(\ref{Za}) ensures that a UAV can offload tasks to a target area only after it has physically reached that area. Constraints~(\ref{Zb}) and~(\ref{Zc}) guarantee that local processing or task offloading is permitted only if the UAV is within the corresponding task area.

Although all decision variables in this formulation are discrete, the overall optimization problem is combinatorially complex and exhibits strong temporal coupling across time slots. As such, the problem constitutes a large-scale, time-dependent INLP problem, which is difficult to solve using conventional optimization techniques in dynamic multi-agent scenarios.

\begin{figure*}[!t]
\centering
\includegraphics[width=15cm]{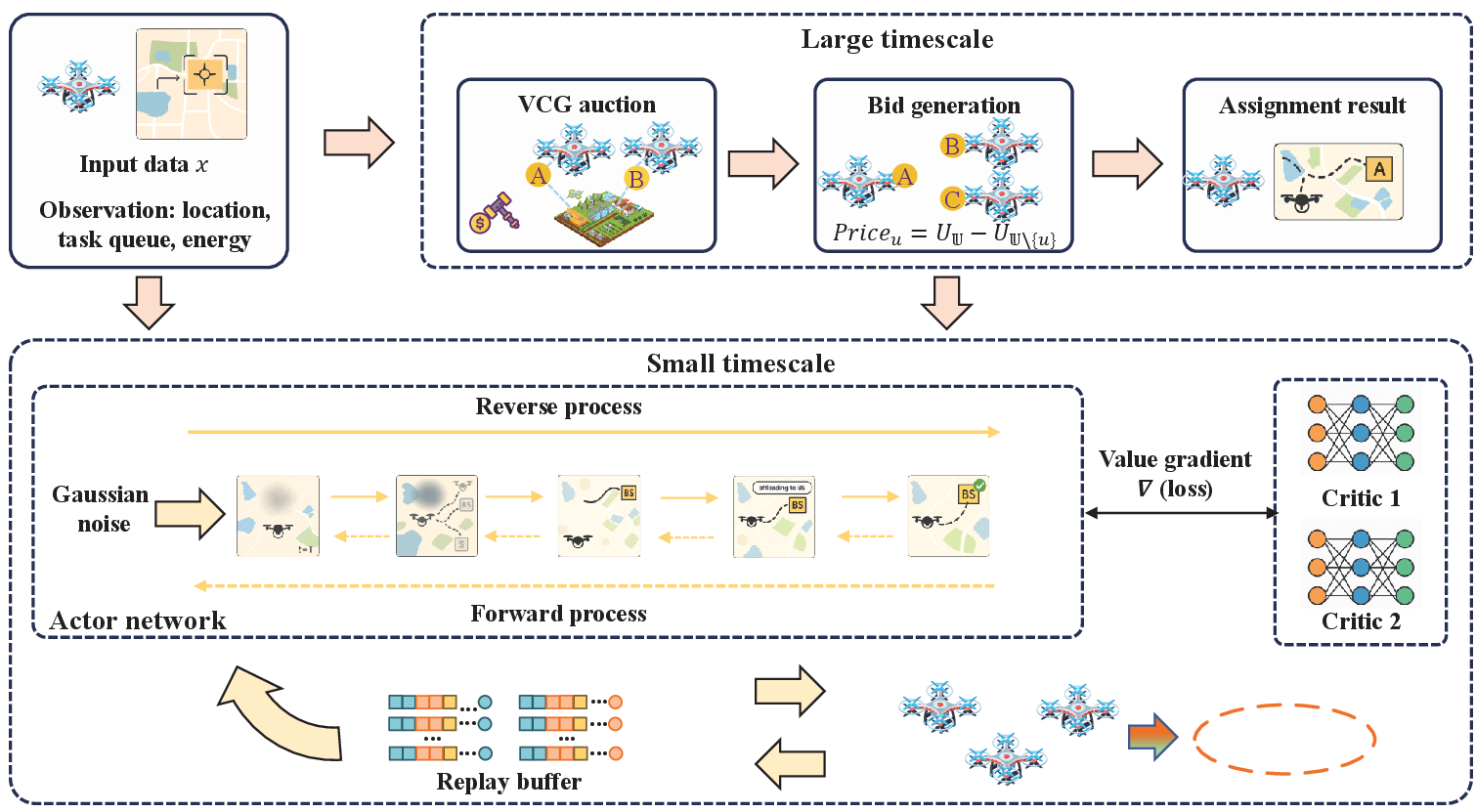}
\caption{Algorithm Design.} 
\label{f3}
\end{figure*}
        
\section{Algorithm Design}\label{s4}  
To address the high-dimensional coupling between trajectory planning and task offloading, we adopt a time-scale separation strategy. This hierarchical design enables tractable optimization by decoupling long-term movement planning and short-term offloading adaptation. UAV trajectories typically evolve over longer intervals due to physical motion constraints, while task arrivals and network conditions fluctuate rapidly. By solving trajectory assignment at a large timescale and task offloading at a small timescale, we balance global coordination with real-time responsiveness.
Therefore, we design the two-stage hierarchical algorithm for optimizing UAV connectivity and task offloading decisions within the IAGCN framework. The optimization process operates across two timescales and includes three key components:

(1) \emph{Task Success and Energy Estimation}: A UAV-side prediction module that estimates the expected energy consumption and task success rate in each task area. These estimates are used to construct utility functions for the bidding process.

(2) \emph{Auction-Based Multi-UAV Task Assignment}: A coordination mechanism that assigns UAVs to task areas using a VCG auction. This mechanism ensures energy-aware and incentive-compatible UAV-task associations.

(3) \emph{D-HAPPO-Based Task Offloading}: A generative MARL algorithm that embeds LDMs into policy networks. This component enables decentralized and adaptive offloading decisions in dynamic environments.

\subsection{Task Success and Energy Consumption Estimation Algorithm}
To evaluate the task success rate and expected energy consumption of a UAV operating in a specific task area, we propose an algorithm estimating the local execution and offloading performance. By comparing the energy required for local processing and offloading to different nodes, the UAV dynamically selects the most energy-efficient execution mode while guaranteeing the task deadlines.

As detailed in Algorithm~\ref{alg:task_success_energy_estimation}, the procedure begins by calculating the horizontal flight distance between the UAV and task area center (line~\ref{a1s1}). Based on the UAV speed, it then computes the travel time and corresponding flight energy consumption (lines~\ref{a1s2}-\ref{a1s3}). 
For each task in the target area, the algorithm estimates the energy required for local processing and compares it with the energy required to offload the task to other nodes (UAVs, TBSs, ABS). The offloading cost consists of communication energy and remote computation energy (lines~\ref{a1s6}-\ref{a1s10}). The execution strategy with the least energy cost is selected (line~\ref{a1s12}). If the task completes before its deadline, it is counted as successful. The total number of successful tasks is used to estimate the task success rate in the area, and the UAV's total energy consumption includes both flight and processing/offloading costs (lines~\ref{a1s17}-\ref{a1s19}).

Let \( N_m \) denote the number of tasks in task area \( m \), and let \( A = |\{\mathbb{U}, \mathbb{B}, \mathbb{H}\}| \) represent the number of candidate offloading nodes (typically a small constant). For each task, the algorithm computes local processing energy once and compares it with offloading energy to all candidate nodes, resulting in \( A \) operations per task. Hence, the total computational complexity per task area is \( \mathcal{O}(AN_m) \). Given that \( A \) is small (e.g., 3), the algorithm scales linearly with the number of tasks and is computationally efficient for online UAV planning.

\begin{algorithm}[!t]
\caption{Task Success and Energy Consumption Estimation}
\label{alg:task_success_energy_estimation}
\begin{algorithmic}[1]
\REQUIRE UAV \( u \), speed \( v_u \), CPU frequency \( f_u \), task area \( m \), and task set \( N_m \).
\ENSURE Estimated task success \( \widetilde{\mathcal{S}}_{m} \), and estimated energy consumption \( \widetilde{\eta}_u(m) \).

\STATE \( D_{u,m} \gets \|l_u(t) - l_m\| \) \label{a1s1}
\STATE \( T_u^{\mathrm{fly}}(m) \gets D_{u,m} / v_u \) \label{a1s2}
\STATE \( E_u^{\mathrm{fly}}(m) \gets P_u^{\mathrm{fly}}(v_u) T_u^{\mathrm{fly}}(m) \) \label{a1s3}
\STATE \( T_{\mathrm{completed}} \gets 0 \)

\FOR{each task \( i \in N_m \)}
    \STATE \( E_{u,i}^{\mathrm{proc}} \gets \kappa_u^{\text{uav}} f_u^2 d_{m,i}  c_{m,i} \) \label{a1s6}
    \FOR{each node \( \alpha \in \{\mathbb{U}, \mathbb{B}, \mathbb{H}\} \)}
        \STATE \( E_{u,i}^{\mathrm{comm}} \gets P_u(t) \dfrac{d_{m,i}}{R_{u,\alpha}^{u2\alpha}(t)} \)
        \STATE \( E_{\alpha,i}^{\mathrm{proc}} \gets \kappa_{\alpha} f_{\alpha}^2 d_{m,i}  c_{m,i} \)
        \STATE \( E_{u,i}^{\mathrm{offload}}(\alpha) \gets E_{u,i}^{\mathrm{comm}} + E_{\alpha,i}^{\mathrm{proc}} \) \label{a1s10}
    \ENDFOR
    \STATE \( E_{u,i} \gets \min\left( E_{u,i}^{\mathrm{proc}}, \min_{\alpha} E_{u,i}^{\mathrm{offload}}(\alpha) \right) \) \label{a1s12}
    \IF{task \( i \) can be completed before deadline}
        \STATE \( T_{\mathrm{completed}} \gets T_{\mathrm{completed}} + 1 \) \label{a1s15}
    \ENDIF
\ENDFOR

\STATE \( \widetilde{\mathcal{S}}_{m} \gets \min\left( T_{\mathrm{completed}}, N_m \right) \) \label{a1s17}
\STATE \( \widetilde{\eta}_u(m) \gets E_u^{\mathrm{fly}}(m) + \sum_{i \in N_m} E_{u,i} \) \label{a1s18}

\RETURN \( \widetilde{\mathcal{S}}_{m},\, \widetilde{\eta}_u(m) \). \label{a1s19}
\end{algorithmic}
\end{algorithm}

\subsection{Auction-Based Multi-UAV Task Assignment Algorithm}\label{alg:auction_based}

To efficiently assign UAVs to different task areas in the IAGCN, we propose an auction-based mechanism that balances energy efficiency and task success. The algorithm is composed of three main phases: utility evaluation and bidding, task area allocation, and VCG-based pricing with trajectory planning. In the bidding phase, each UAV \( u \in \mathbb{U} \) independently evaluates the utility of each task area \( m \in \mathcal{D} \) by jointly considering its predicted energy expenditure and the task completion rate in that area. The estimated utility is 
\begin{equation}
U_u(m) = \gamma_1 \widetilde{\eta}_u(m) + \gamma_2 \widetilde{\mathcal{S}}_{m},
\end{equation}
where \( \widetilde{\eta}_u(m) \) denotes the estimated energy consumption for UAV \( u \) in area \( m \), and \( \widetilde{\mathcal{S}}_{m} \) represents the predicted number of successfully completed tasks. The weights \( \gamma_1 \) and \( \gamma_2 \) balance the energy efficiency and task success probability.

Based on its utility estimate, each UAV submits a bid:
\begin{equation}
b_u(m) = \beta U_u(m),
\end{equation}
where \( \beta \) is a scaling coefficient to ensure the numerical stability during comparison. All bids are submitted to the auctioneer (e.g., ABS) for centralized evaluation.

Once bids are received, the auctioneer begins assigning UAVs to task areas iteratively. Each UAV is provisionally assigned to the task area for which it submits the highest bid. Task areas maintain a candidate assignment set \( S[m] \), and the price for each area is updated based on aggregated bids.

The allocation proceeds in rounds, where unassigned UAVs are iteratively matched to their preferred areas. If multiple UAVs bid for the same area, tie-breaking rules (e.g., earliest bid or smallest estimated cost) are used. The process terminates when all UAVs are assigned or no further feasible assignments can be made.

\begin{algorithm}[!t]
\caption{Auction-Based Multi-UAV Task Assignment}
\label{alg:auction_based}
\begin{algorithmic}[1]
\REQUIRE UAV set \( \mathbb{U} \), task area set \( \mathcal{D} \), estimated success \( \widetilde{\mathcal{S}}_{m} \), energy \( \widetilde{\eta}_u(m) \), and parameters \( \gamma_1, \gamma_2, \beta \).
\ENSURE Assigned task areas \( \{a_u\}_{u \in \mathbb{U}} \) and VCG prices \( \{\text{Price}_u\}_{u \in \mathbb{U}} \).

\FOR{each UAV \( u \in \mathbb{U} \)}
    \FOR{each task area \( m \in \mathcal{D} \)}
        \STATE Compute utility \( U_u(m) = \gamma_1 \widetilde{\eta}_u(m) + \gamma_2  \widetilde{\mathcal{S}}_{m} \).
        \STATE Set bid \( b_u(m) = \beta  U_u(m) \).
    \ENDFOR
    \STATE Submit bid vector \( \mathbf{b}_u \).
\ENDFOR

\STATE Initialize assignment set \( S[m] \gets \emptyset \) for all \( m \in \mathcal{D} \).

\FOR{each UAV \( u \)}
    \STATE Select preferred task area \( m^* = \arg\max b_u(m) \).
    \STATE Assign \( u \to m^* \), update \( S[m^*] \), and store utility \( U_u(m^*) \).
\ENDFOR

\FOR{each UAV \( u \)}
    \STATE Compute system utility \( U_{\mathbb{U}} \).
    \STATE Remove \( u \), compute \( U_{\mathbb{U} \setminus \{u\}} \).
    \STATE Set \( \text{Price}_u = U_{\mathbb{U}} - U_{\mathbb{U} \setminus \{u\}} \).
    \STATE Plan trajectory to assigned area \( {a_u} \).
\ENDFOR

\RETURN \( \{a_u\} \), and \( \{\text{Price}_u\} \).

\end{algorithmic}
\end{algorithm}

After finalizing the UAV-task assignments, the auctioneer computes a payment for each UAV using the VCG mechanism. 
This pricing strategy ensures that UAVs minimize their payments by reporting their true utilities, thus encouraging truthful bidding and aligning individual incentives with global efficiency.
The price reflects the marginal contribution of UAV \( u \) to the total utility and is defined as
\begin{equation}
\text{Price}_u = U_{\mathbb{U}} - U_{\mathbb{U} \setminus \{u\}},
\end{equation}
where \( U_{\mathbb{U}} \) is the total utility with all UAVs present, and \( U_{\mathbb{U} \setminus \{u\}} \) is the utility if UAV \( u \) is removed from the assignment. This pricing strategy promotes truthfulness by ensuring that each UAV minimizes its cost by bidding its true utility.

Upon assignment, each UAV plans a flight path to its designated task area. The trajectory is optimized to minimize flight energy consumption \( \widetilde{\eta}_u(a_u) \), based on the current location and spatial constraints. Since the flight energy dominates overall energy usage, careful planning ensures efficient UAV repositioning and timely task initiation.

Algorithm~\ref{alg:auction_based} presents the overall procedure for the auction-based multi-UAV task assignment. The algorithm begins with the initialization of assignment states and utility evaluations, followed by UAVs computing their bids for each task area and submitting them to the auctioneer (lines~1-7). In the next phase, an iterative allocation mechanism assigns UAVs to task areas based on the highest submitted bids, ensuring the fair and efficient distribution (lines~8-12). After the task assignments are finalized, the VCG mechanism is applied to determine the marginal pricing for each UAV, ensuring incentive compatibility (lines~13-16). Finally, each UAV plans an energy-efficient trajectory to its assigned area (line 17), and the algorithm returns the assignments and prices (line 19).

Let \( U = |\mathbb{U}| \) be the number of UAVs and \( M = |\mathcal{D}| \) be the number of task areas. The bid computation phase requires \( \mathcal{O}(UM) \) operations. In the worst case, the iterative allocation process assigns each UAV by taking an \(\arg\max\) over \(M\) areas, which runs in \( \mathcal{O}(UM) \) without per-UAV sorting.
The VCG pricing phase requires evaluating marginal utility differences for each UAV, which can be computed with precomputed totals in \( \mathcal{O}(U) \).
Overall, the algorithm has a total time complexity of \( \mathcal{O}(UM) \), making it scalable for moderate swarm sizes in practical low-altitude network settings.

\begin{algorithm}[!t]
\caption{D-HAPPO Training Algorithm}
\label{alg:dhappo}
\begin{algorithmic}[1]
\REQUIRE Stepsize \( \alpha \), batch size \( B \), number of agents \( n \), total episodes \( K \), and steps per episode \( T \).
\STATE Initialize actor parameters \( \{ \theta^i \}_{i=1}^n \), critic parameters \( \psi \), and buffer \( \mathcal{B} \).
\FOR{episode \( k = 1 \) to \( K \)}
    \STATE Collect trajectories \( (s_t, o_t^i, a_t^i, r_t, s_{t+1}) \).
    \STATE Store transitions in replay buffer \( \mathcal{B} \).
    \STATE Sample minibatch \( \mathcal{D} \subset \mathcal{B} \), compute generalized advantage estimation \( \hat{A}(s_t, \bm{a}_t) \).
    \STATE Generate permutation \( \pi = [i_1, \dots, i_n] \).
    \STATE Initialize \( M^{i_1}(s_t, \bm{a}_t) = \hat{A}(s_t, \bm{a}_t) \).
    \FOR{each agent \( i_m \in \pi \)}
        \STATE Encode \( o_t^{i_m} \to f_t \), sample noise \( \epsilon \).
        \STATE Construct \( \mathbf{z}_t = \sqrt{\bar{\alpha}_t} \mathbf{z}_0 + \sqrt{1 - \bar{\alpha}_t} \epsilon \).
        \STATE Denoise \( \mathbf{z}_t \to \hat{\mathbf{z}}_0 \), decode to action \( a_t^{i_m} \).
        \STATE Compute loss \( \mathcal{L}^{i_m} \), update \( \theta^{i_m} \leftarrow \theta^{i_m} - \alpha \nabla \mathcal{L}^{i_m} \).
        \STATE Update compound advantage \( M^{i_{1:m+1}} \).
    \ENDFOR
    \STATE Update critic parameters \( \psi \) using TD loss.
\ENDFOR
\end{algorithmic}
\end{algorithm}

\subsection{D-HAPPO for Task Offloading}
After the VCG-based auction determines the large timescale assignment of UAVs to task areas, each UAV must make fine-grained decisions regarding task execution within its assigned region, such as selecting offloading targets, adapting to local dynamics, and navigating environmental uncertainties. To address these small timescale decisions in a decentralized and adaptive manner, we propose the D-HAPPO framework, which integrates LDMs into a generative MARL structure to enhance policy diversity and adaptability under uncertainty.

To capture the decentralized and dynamic decision-making process, we formulate the problem as a Markov game.
In this setting, UAVs are modeled as $N$ cooperative agents interacting in a shared environment, and the task offloading process is formally represented as a Markov game defined by the tuple \( \langle \mathcal{S}, \mathcal{O}, \mathcal{D}, \mathcal{R}, \gamma \rangle \), detailed as follows. 

\paragraph{State Space}
Each agent \( n \in \mathbb{U} \) has a state \( s_n(t) \) that includes its own position \( l_u(t) \), the positions of TBSs and ABS \( l_b(t), l_h(t) \), task area information \( \mathcal{D}_m^{\text{ta}} \), and task queues \( \mathcal{Q}^{\text{ta}} and \mathcal{Q}^{\text{uav}} \). Formally,
\begin{equation}
s_n(t) = \{l_u(t), l_b(t), l_h(t), \mathcal{D}_m^{\text{ta}}, \mathcal{Q}^{\text{ta}}, \mathcal{Q}^{\text{uav}}\}.
\end{equation}

\paragraph{Observation Space}
Each agent only observes local and partially shared information. The observation space \( \mathcal{O}_n \subseteq \mathcal{S} \) includes:
\begin{equation}
    o_n(t) = \{l_u(t), l_b(t), l_h(t), \mathcal{D}_m^{\text{ta}}, \mathcal{Q}^{\text{ta}}, \mathcal{Q}^{\text{uav}}\}.
\end{equation}

\paragraph{Action Space}
The action space \( \mathcal{D}_n \) includes all possible offloading destinations for the task currently assigned to UAV \( n \). The agent chooses to either process locally or offload to a candidate node \( x \in \mathbb{U} \cup \mathbb{B} \cup \mathbb{H} \):
\begin{equation}
    \mathcal{D}_n = \{ x \mid x \in \mathbb{U} \cup \mathbb{B} \cup \mathbb{H} \}.
\end{equation}

\paragraph{Reward Function}
The reward function \( \mathcal{R}_n \) guides each UAV to balance the energy cost, timeliness, and task completion:
\begin{equation}
    \mathcal{R}_n = -\alpha E_n + \gamma S_n,
\end{equation}
where \( E_n \) is the total energy consumed for processing or offloading the task, and \( S_n \in \{0,1\} \) indicates task success. The parameter \( \alpha \) penalizes energy usage, and \( \gamma \) rewards successful completion.

As shown in Fig. \ref{f3}, the forward diffusion process progressively perturbs a clean latent vector by adding Gaussian noise over multiple steps, gradually transforming it into a standard normal distribution. Conversely, a randomly sampled latent variable is denoised by the reverse diffusion process, conditioned on the agent's observation and time step, to reconstruct a meaningful latent representation. This generative trajectory enables the policy to flexibly reconstruct diverse actions from noisy priors.
Specifically, we introduce a standard Gaussian latent variable \( \mathbf{z}_0 \sim \mathcal{N}(0, I) \), which serves as a learnable prior for action generation. The forward diffusion process progressively injects noise into \( \mathbf{z}_0 \) over \( T \) steps using a variance schedule \( \bar{\alpha}_t \), yielding a noisy latent variable:
\begin{equation}
\mathbf{z}_t = \sqrt{\bar{\alpha}_t} \mathbf{z}_0 + \sqrt{1 - \bar{\alpha}_t} \epsilon, \quad \epsilon \sim \mathcal{N}(0, I).
\end{equation}

The reverse process is modeled by a denoising network \( D_\theta \), which attempts to recover \( \epsilon \) from \( \mathbf{z}_t \), conditioned on the observation embedding \( \phi(o_n) \) and timestep \( t \). The training objective is:
\begin{equation}
\mathcal{L}_{\text{diff}} = \mathbb{E}_{t, \mathbf{z}_0, \epsilon} \left[ \left\| \epsilon - D_\theta(\mathbf{z}_t, \phi(o_n), t) \right\|^2 \right].
\end{equation}

During the inference, a clean latent vector \( \hat{\mathbf{z}}_0 \) is obtained by reverse sampling from a random noise input \( \mathbf{z}_t \), using the trained denoiser:
\begin{equation}
\hat{\mathbf{z}}_0 = D_\theta(\mathbf{z}_t, \phi(o_n), t).
\end{equation}

The latent is then decoded into a final action \( a_n \) via a deterministic mapping \( \psi \), expressed as \( a_n = \psi(\hat{\mathbf{z}}_0) \). This conditional generative process introduces a rich, observation-driven action space that enhances the agent's adaptability to dynamic environments.
Unlike standard Gaussian noise-based exploration, the diffusion actor network generates actions by denoising a latent prior conditioned on the observation, enabling more diverse, robust, and observation-adaptive policy behaviors under uncertainty.

To ensure the cooperative convergence, we embed the LDM-based actor into the sequential update structure of HAPPO~\cite{HARL_2024}. For agent \( i_m \), the clipped HAPPO loss is computed as:
\begin{equation}
\begin{split}
\mathcal{L}(\theta_{k+1}^{i_m}) = &\mathbb{E}_{s, \bm{a}} \Bigg[ \min \Bigg( \frac{\pi_{i_m}^{\theta_{i_m}}(a_{i_m}|s)}{\pi_{i_m}^{\theta_{i_m}^k}(a_{i_m}|s)} M^{i_{1:m}}(s, \bm{a}), \\
&\text{clip}\left( \frac{\pi_{i_m}^{\theta_{i_m}}(a_{i_m}|s)}{\pi_{i_m}^{\theta_{i_m}^k}(a_{i_m}|s)}, 1 \pm \epsilon \right) M^{i_{1:m}}(s, \bm{a}) \Bigg) \Bigg],
\end{split}
\end{equation}
where \( \pi^{\theta^k} \) is the old policy, and \( M^{i_{1:m}} \) is the recursively computed compound advantage:
\begin{equation}
M^{i_{1:m}}(s, \bm{a}) = \frac{\pi^{i_{1:m-1}}_{\theta_{k+1}}(\bm{a}^{i_{1:m-1}}|s)}{\pi^{i_{1:m-1}}_{\theta_k}(\bm{a}^{i_{1:m-1}}|s)} \hat{A}(s, \bm{a}).
\end{equation}

The structure ensures each agent is trained with consistent information, while allowing latent diffusion to shape its own stochastic policy representation. D-HAPPO provides a unified framework that leverages the strengths of generative modeling and multi-agent actor-critic learning, offering a scalable solution to real-world UAV coordination in edge intelligence systems.

\section{Simulation Results}\label{s5}

To evaluate the performance of the proposed framework in hierarchical UAV task offloading, we conduct extensive simulations under realistic low-altitude edge computing scenarios. The effectiveness of our approach is benchmarked with four baselines.

\subsubsection{Greedy-Match and Shortest-Path (GM-SP)}
This baseline assigns each UAV to the geographically nearest task area and assumes local task execution via the shortest route. It reflects a simple greedy strategy that prioritizes distance minimization without considering resource constraints or task deadlines.

\subsubsection{Max-Utility Matching and Static Offloading (MU-SO)}
In this rule-based approach, UAVs are assigned to task areas based on a weighted utility score incorporating proximity and task urgency. Offloading decisions follow static rules: UAVs offload to TBSs or the ABS based on fixed thresholds, without adaptation to queue states or channel conditions.

\subsubsection{Load-Balanced Rule-Based Offloading (LB-RBO)}
Tasks are distributed among available nodes based on queue length and remaining computing resources. This method aims to balance load across UAVs, TBSs, and the ABS to reduce task latency, but lacks learning capability and adaptability.

\subsubsection{HAPPO}
HAPPO~\cite{HARL_2024} represents a state-of-the-art MARL algorithm. It supports decentralized actor updates with a shared critic, enabling coordinated policy learning in heterogeneous UAV environments.

\begin{table}[!t]
\caption{Simulation Parameters}
\centering
\begin{tabular}{|m{1.7cm}<{\centering}|m{1.8cm}<{\centering}||m{1.6cm}<{\centering}|m{1.8cm}<{\centering}|}
\hline
\textbf{Parameter} & \textbf{Value} & \textbf{Parameter} & \textbf{Value} \\ \hline
    $f_c$ & 2.4 GHz &  $N_G$ & $-114$ dBm \\ \hline
$B$ & 10 MHz &  $H$ & 30 m \\ \hline
$v_u^{\max}$ & 15 m/s & $P_u^{\mathrm{fly}}$ & 35 W \\ \hline
$p_u^{\mathrm{stop}}$ & 10 W & $p_u^{\mathrm{max}}$ & 60 W \\ \hline
$d_{m,n}$ & [0.5,1.0] Mbits & $c_{m,n}$ & 300 cycles/bit \\ \hline
$f_u$, $f_b$, $f_h$ & 2, 3, 4 GHz & $\kappa_u^{\text{uav}}$ & $10^{-28}$ \\ \hline
Learning rate & $1 \times 10^{-4}$ & Batch size & 128 \\ \hline
\end{tabular}
\label{T2}
\end{table}

\begin{figure*}[htbp]
\centering
\subfloat[Reward with different activation functions.]{
\includegraphics[width=7cm]{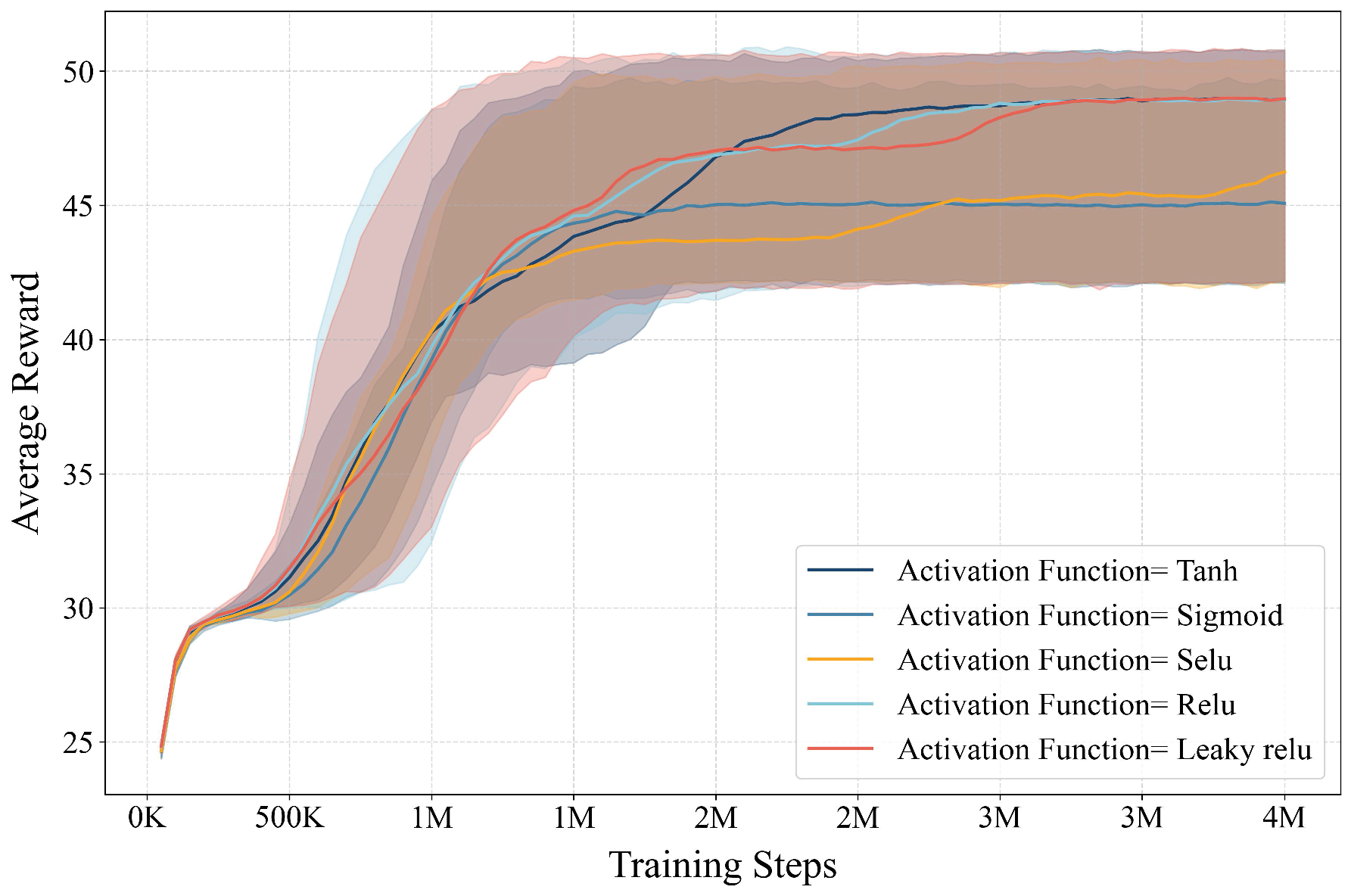}\label{f4a}
}
\subfloat[Reward with different actor learning rates.]{
\includegraphics[width=7cm]{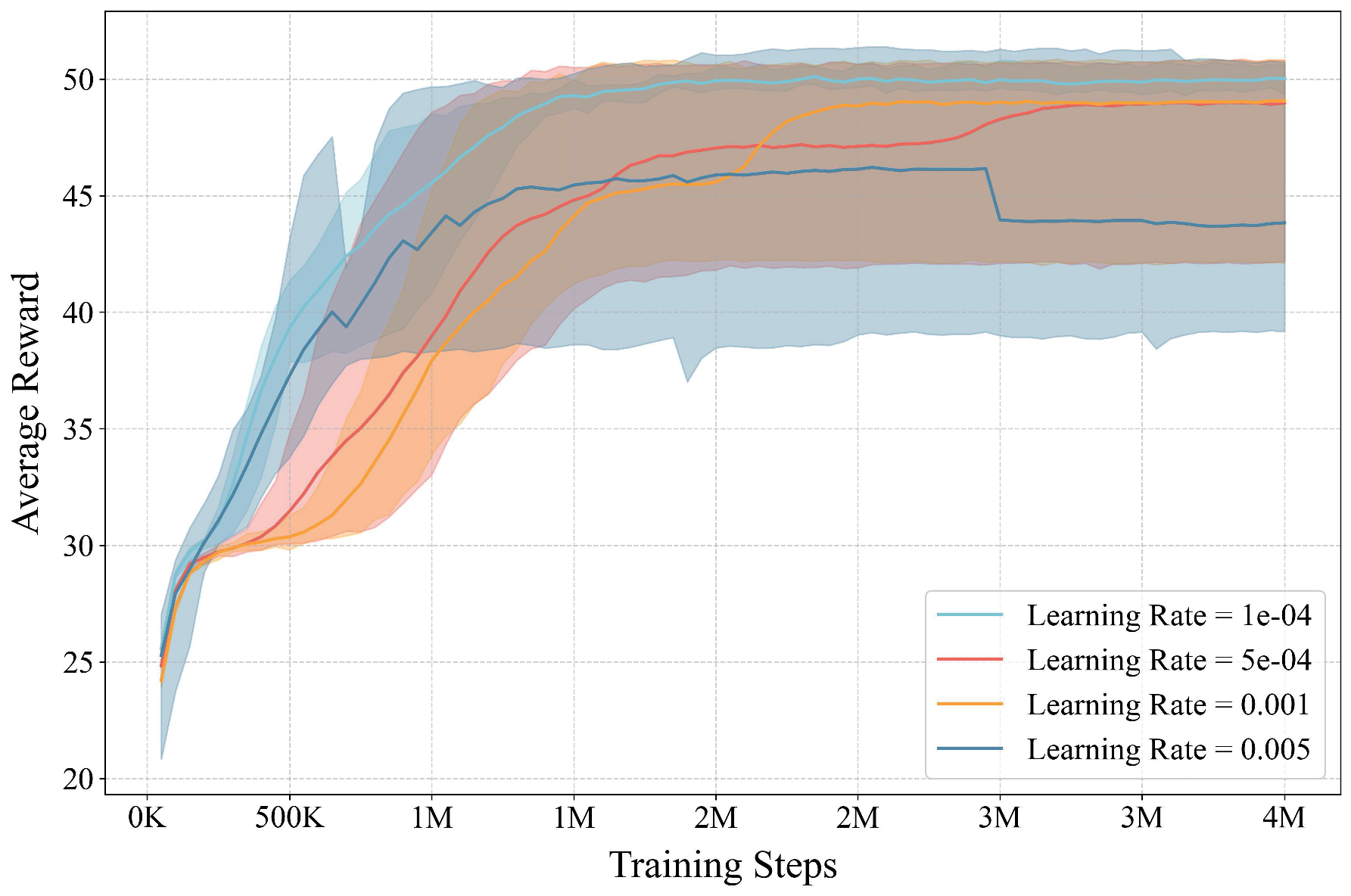}\label{f4b}
}\\
\subfloat[Reward with different hidden sizes.]{
\includegraphics[width=7cm]{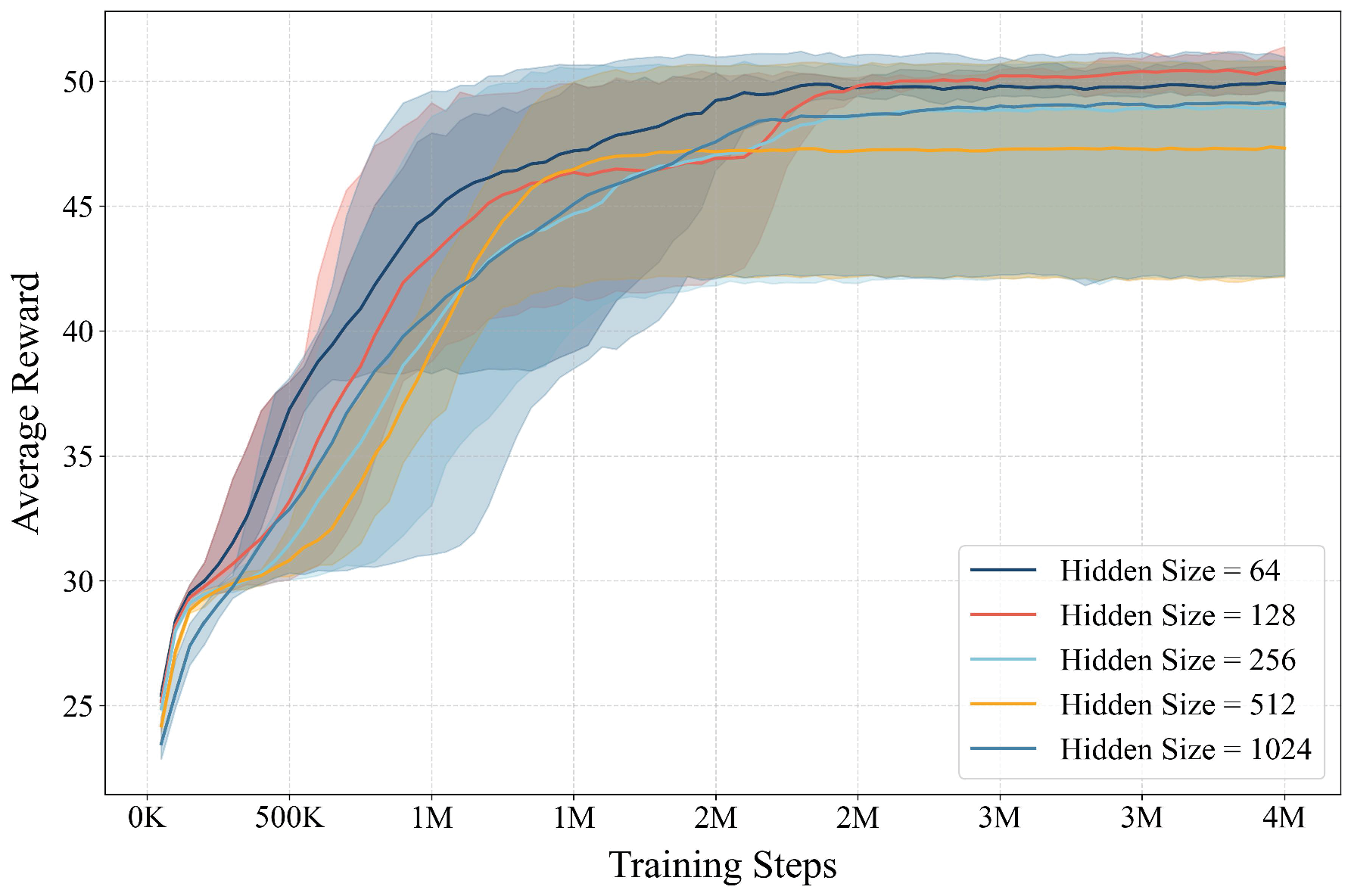}\label{f4c}          
}
\subfloat[Reward with different latent dimensions.]{
\includegraphics[width=7cm]{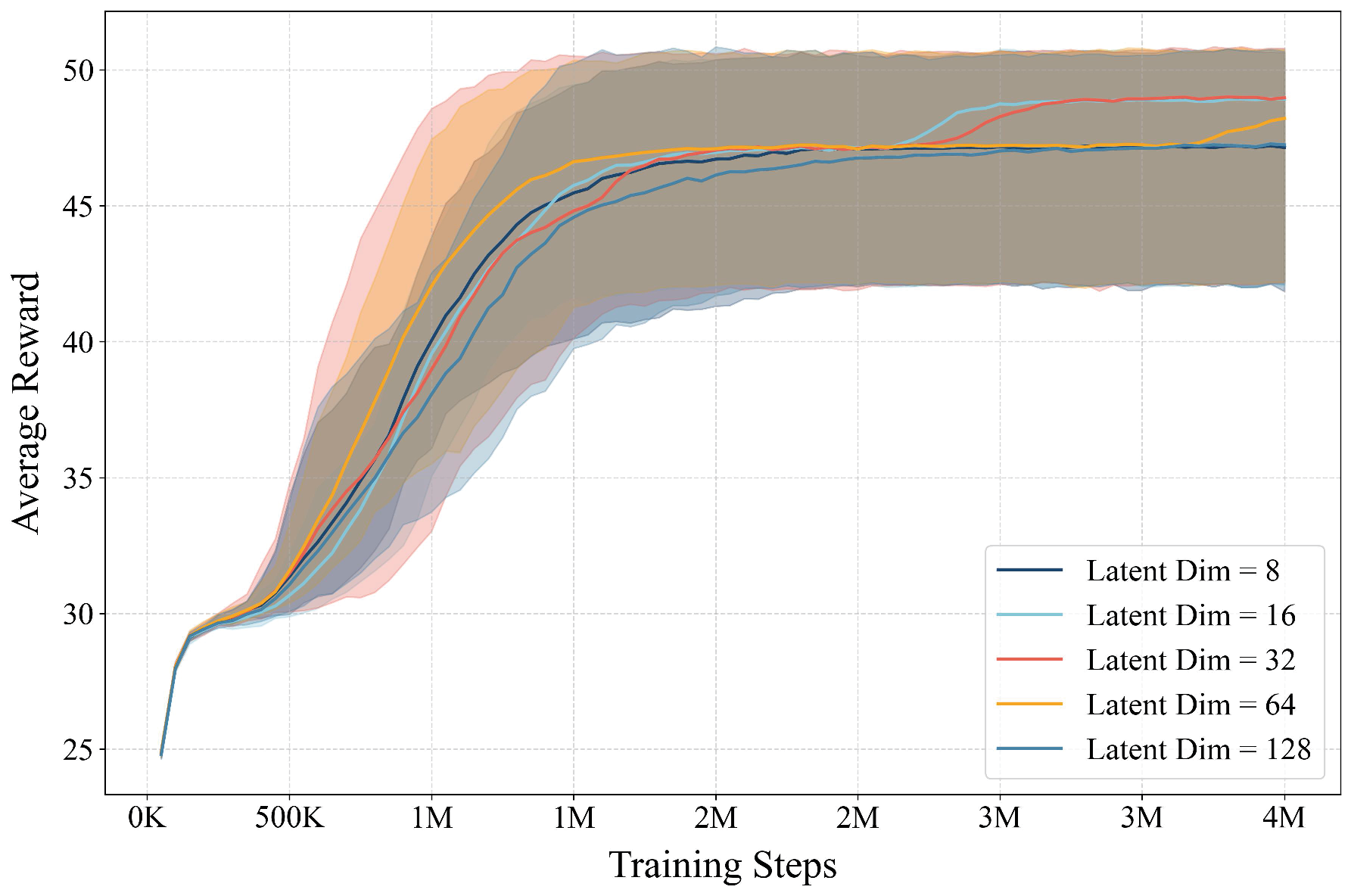}\label{f4d}
}
\caption{Performance with different parameters.}
\label{f4}
\end{figure*}

\subsection{Simulation Setup}
We simulate a \(1,000\,\text{m} \times 1,000\,\text{m}\) low-altitude edge computing environment, consisting of 4 predefined task areas \cite{JCOR_2025}. 
We assume that task areas are fixed and known a priori, which aligns with practical scenarios such as disaster monitoring or infrastructure inspection, where such regions are typically predefined.
Each area maintains an initialized queue of computation-intensive tasks, with task parameters such as data size and deadline drawn randomly at the beginning of the episode. 

Six UAVs are deployed to traverse these areas, collaboratively completing tasks. Additionally, two TBSs  and one ABS are available to support task offloading. 
To ensure fair comparison and isolate algorithmic effects, all UAVs are configured with the same CPU frequency, storage capacity, and communication range during simulation.
At the large timescale, UAVs are assigned to task areas using an auction-based coordination mechanism, while at the small timescale, each UAV dynamically selects whether to process tasks locally or offload them to nearby nodes based on the current network and resource conditions.
Each simulation episode spans \(T = 100\) time slots, during which agent mobility, task scheduling, communication, and policy learning evolve in real time. All agents operate under practical constraints on bandwidth, computation capacity, and transmission power. Key simulation parameters are listed in Table~\ref{T2}.

\subsection{Performance with Different Parameters}

To understand the sensitivity and robustness of D-HAPPO under different architectural and optimization settings, we conduct controlled experiments by varying four key parameters: activation functions, learning rates, hidden sizes, and latent dimensions. Fig.~\ref{f4} presents the training curves in terms of the average reward across 4 million steps.

The activation function directly influences the non-linearity and gradient propagation properties of neural networks. As illustrated in Fig.~\ref{f4}(a), Tanh and ReLU deliver superior performance, reaching average rewards close to 50. Tanh provides smoother gradient flow due to its bounded nature, which aids convergence stability in early phases. ReLU converges slightly faster but exhibits higher variance, especially after 2M steps, due to unbounded outputs leading to occasional value explosions. Sigmoid performs poorly in both convergence rate and final performance, due to gradient saturation issues. Interestingly, Selu and Leaky ReLU yield more stable but conservative policies, which benefit safety-critical applications where performance oscillation is undesirable.

The learning rate is critical for balancing exploration and convergence. As shown in Fig.~\ref{f4}(b), a learning rate of $5\times10^{-4}$ offers the best trade-off, enabling fast convergence and final performance. The highest rate tested (0.001) accelerates initial learning but becomes unstable in the long term, suggesting overshooting of local minima in actor updates. Conversely, smaller rates such as $1\times10^{-4}$ lead to slow but steady improvements, which might be preferable in environments with highly non-stationary dynamics. The performance degradation of the $0.005$ curve further confirms that overly aggressive updates harm the policy stability in multi-agent coordinations.

The hidden size determines the representational capacity of policy and value networks. As shown in Fig.~\ref{f4}(c), networks with hidden sizes of 128 and 256 achieve the best trade-off between convergence speed and final reward, demonstrating strong performance and stable training. The model with size 64 converges early but to a lower reward ceiling, indicating underfitting due to insufficient expressiveness. Larger models, such as size 512 and 1024, exhibit slower convergence and higher variance in the early stages. Notably, the model with size 1,024 shows instability and wider confidence bounds, caused by over-parameterization and gradient noise. 

The latent dimension governs the representational richness of the generative policy. As illustrated in Fig.~\ref{f4}(d), all latent dimensions achieve comparable final rewards, indicating that the model is relatively robust to this hyperparameter. Smaller dimensions (e.g., 8 and 16) converge steadily but slightly lag behind in early performance, suggesting limited expressiveness during exploration. Moderate dimensions such as 32 and 64 provide faster convergence and more stable reward curves, benefiting from better stochasticity and flexibility. However, increasing the dimension to 128 has no significant performance gain and introduces slightly wider confidence bounds, due to the increased sampling variance. A higher-dimensional latent space increases the variance of the Gaussian noise sampled during the forward diffusion process, which may lead to greater uncertainty in the reconstructed actions and reduced stability during early training.

\begin{figure*}[htbp]
\centering
\subfloat[Energy efficiency under different UAV numbers.]{
\includegraphics[width=5.8cm]{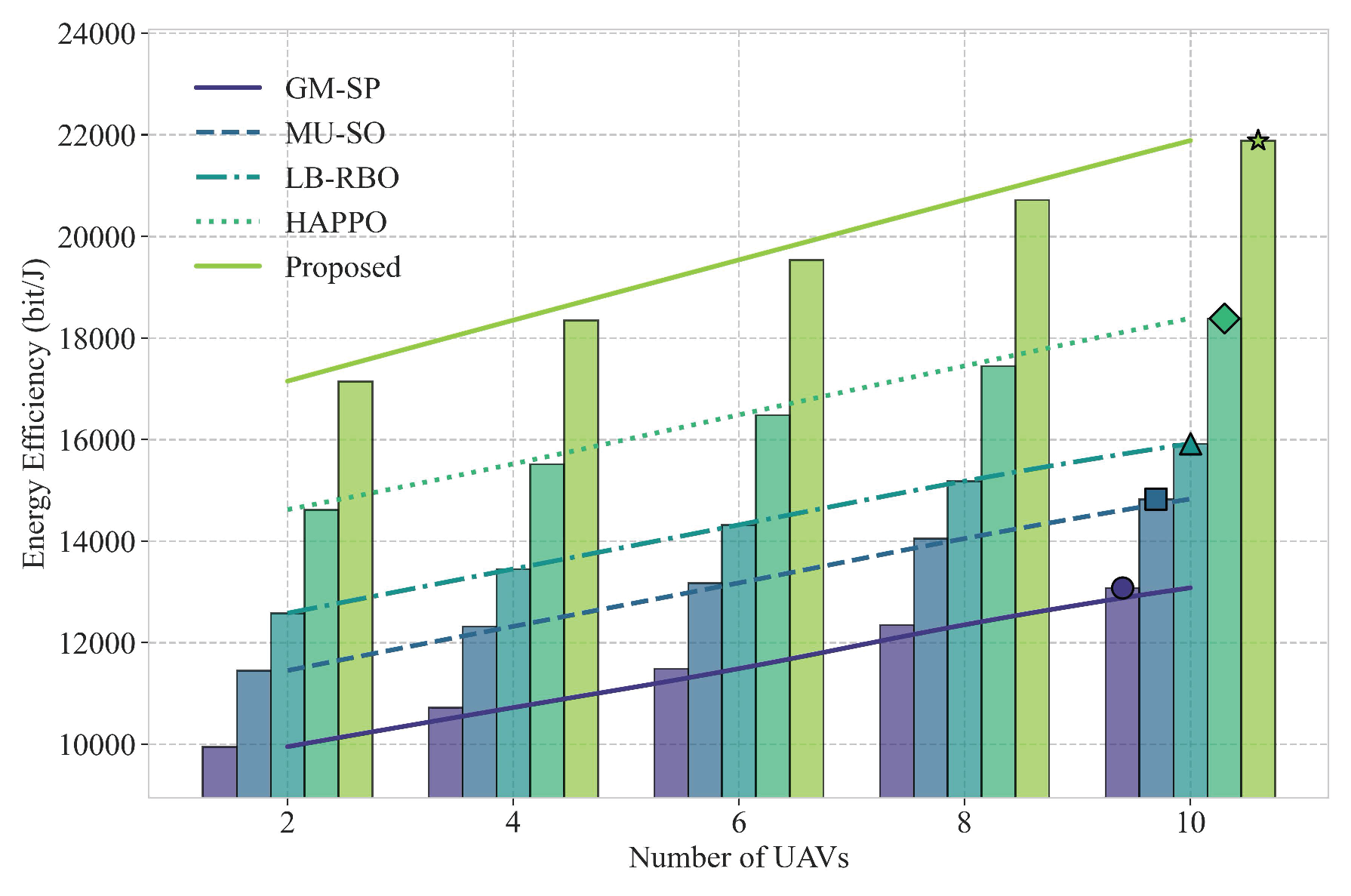}}
\hfill
\subfloat[Task completion ratio under different UAV numbers.]{
\includegraphics[width=5.8cm]{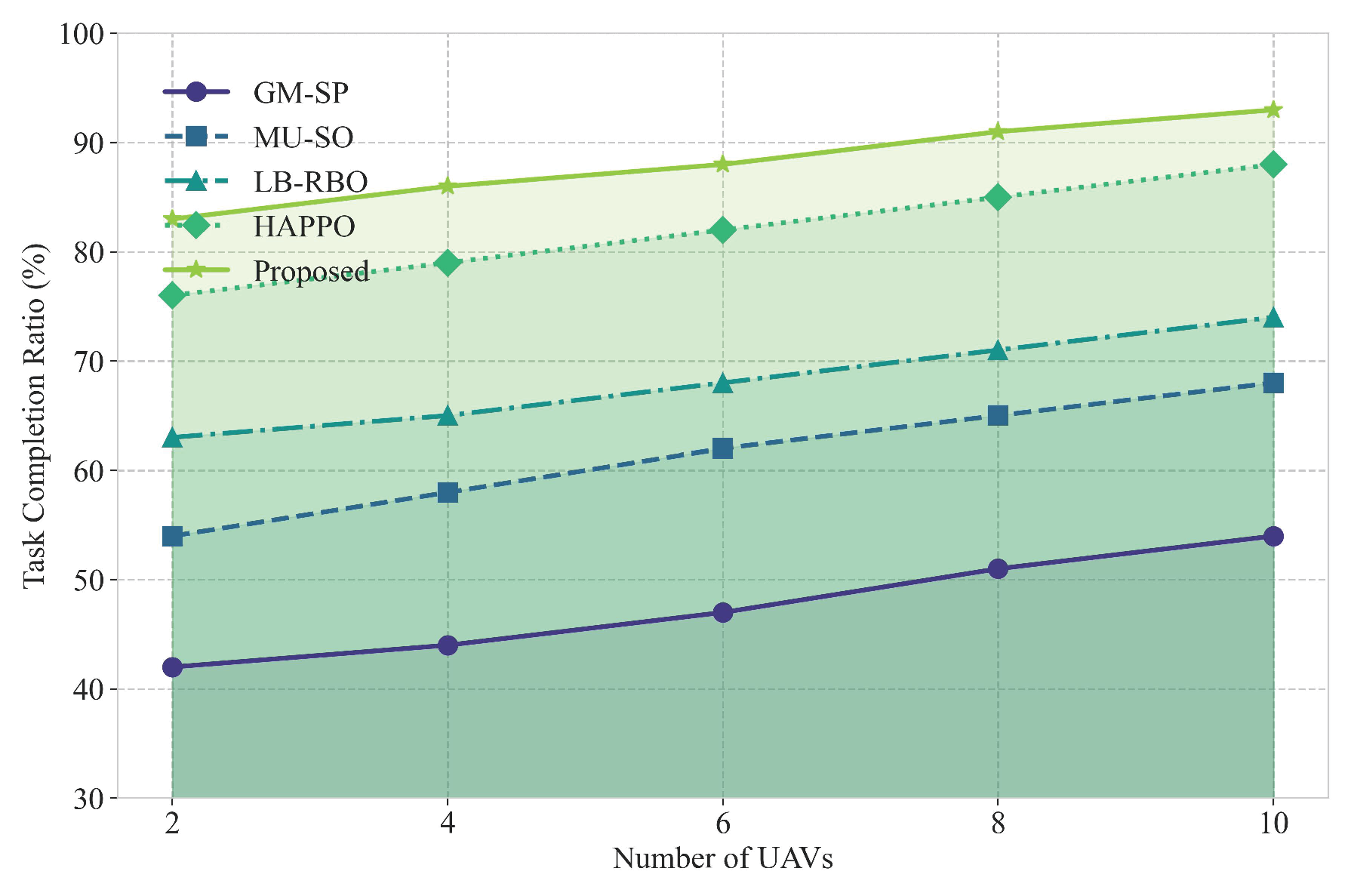}}
\hfill
\subfloat[Average task latency under different UAV numbers.]{
\includegraphics[width=5.8cm]{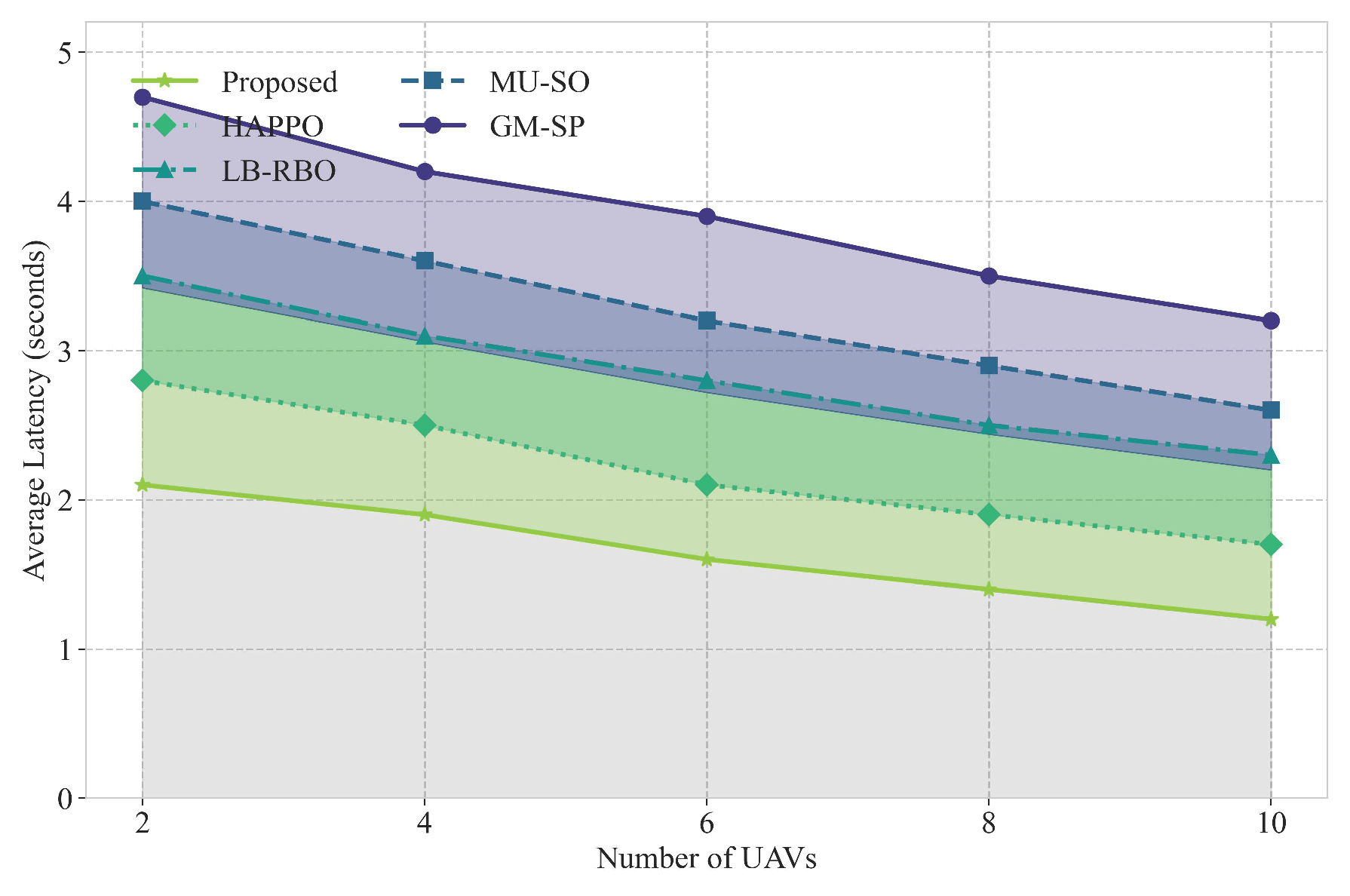}}
\caption{System performance under varying UAV numbers.}
\label{f5}
\end{figure*}

\begin{figure*}[htbp]
\centering
\subfloat[Energy efficiency under different task sizes.]{
\includegraphics[width=5.8cm]{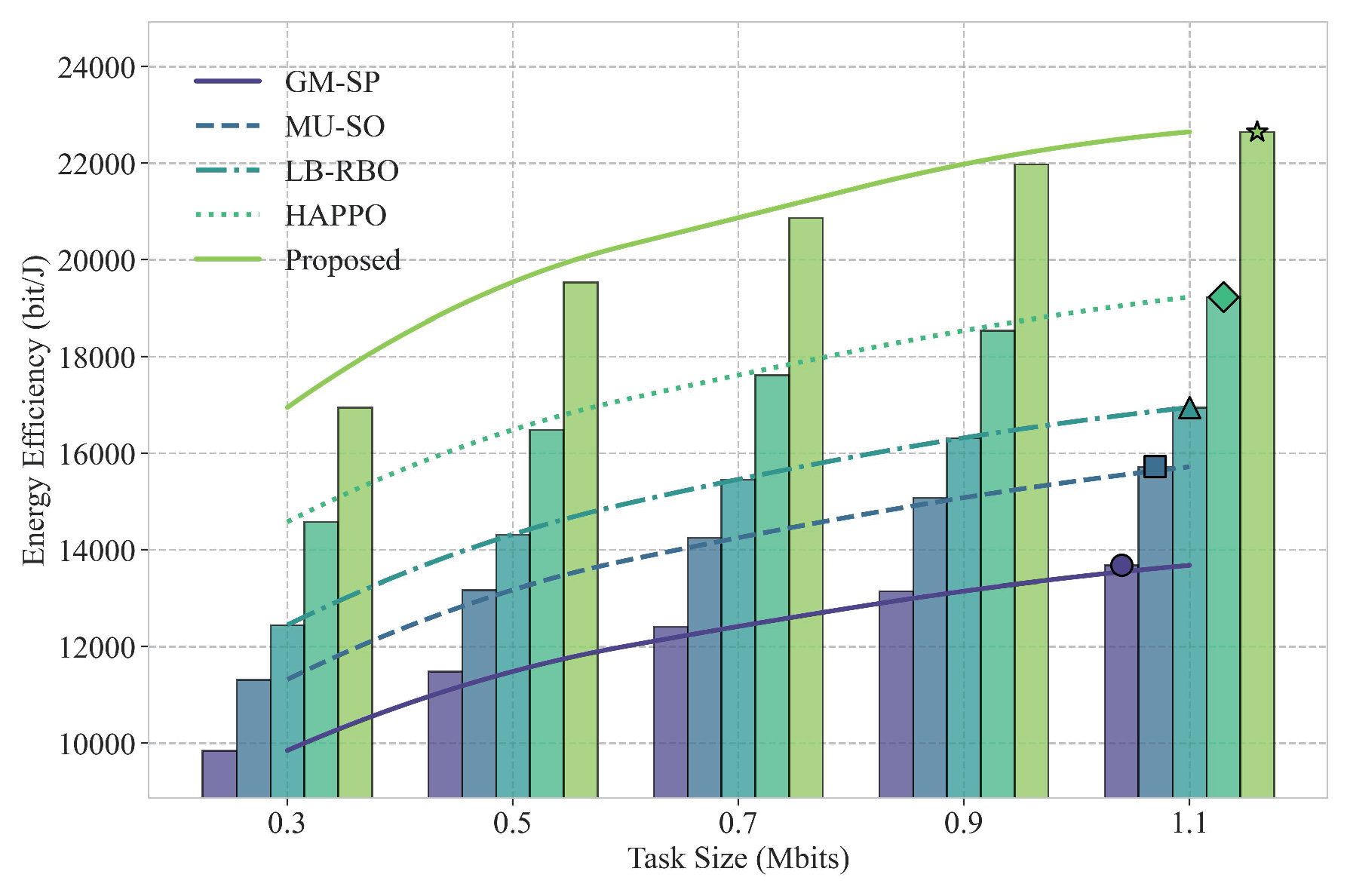}}
\hfill
\subfloat[Task completion ratio under different task sizes.]{
\includegraphics[width=5.8cm]{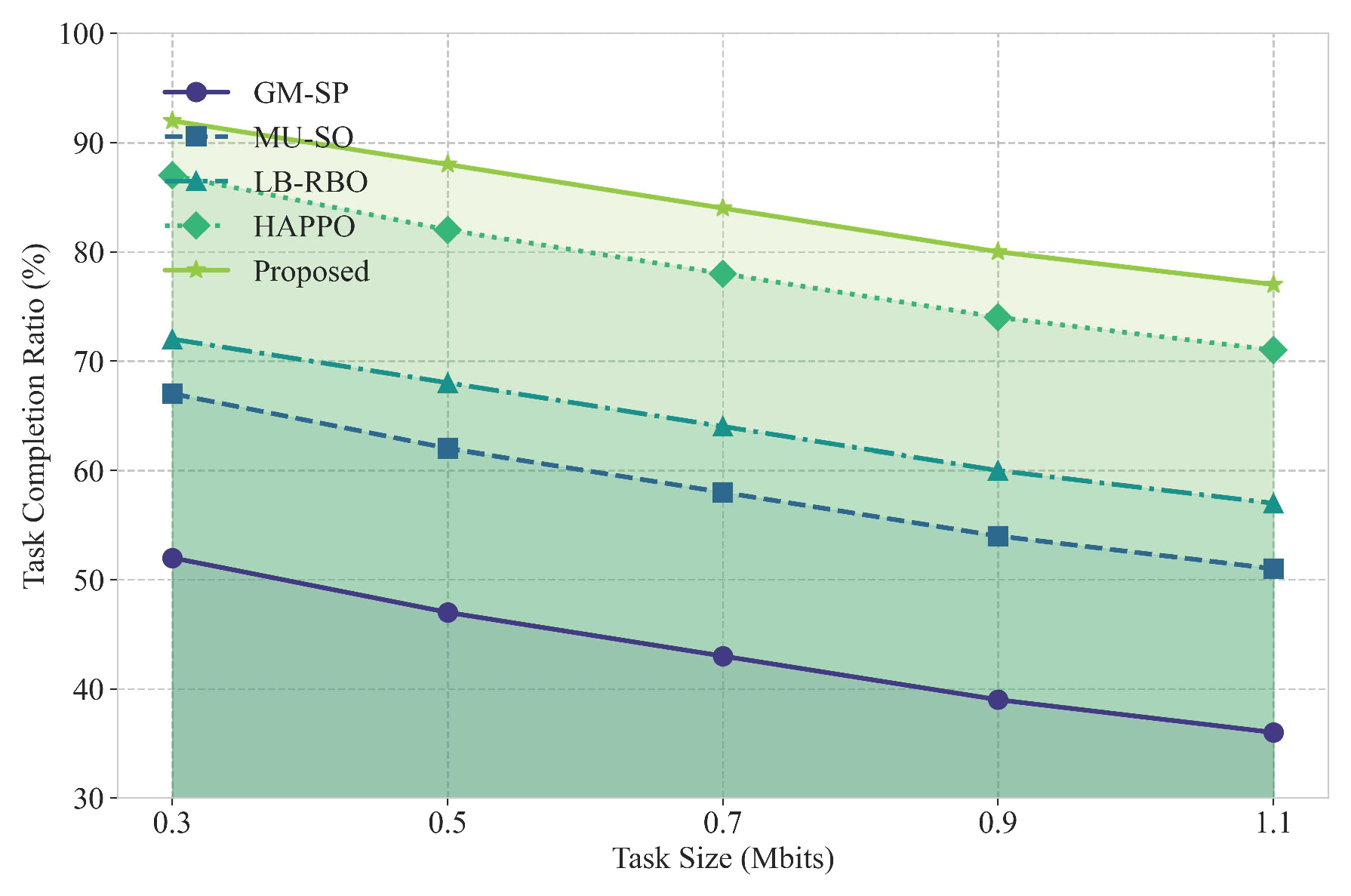}}
\hfill
\subfloat[Average task latency under different task sizes.]{
\includegraphics[width=5.8cm]{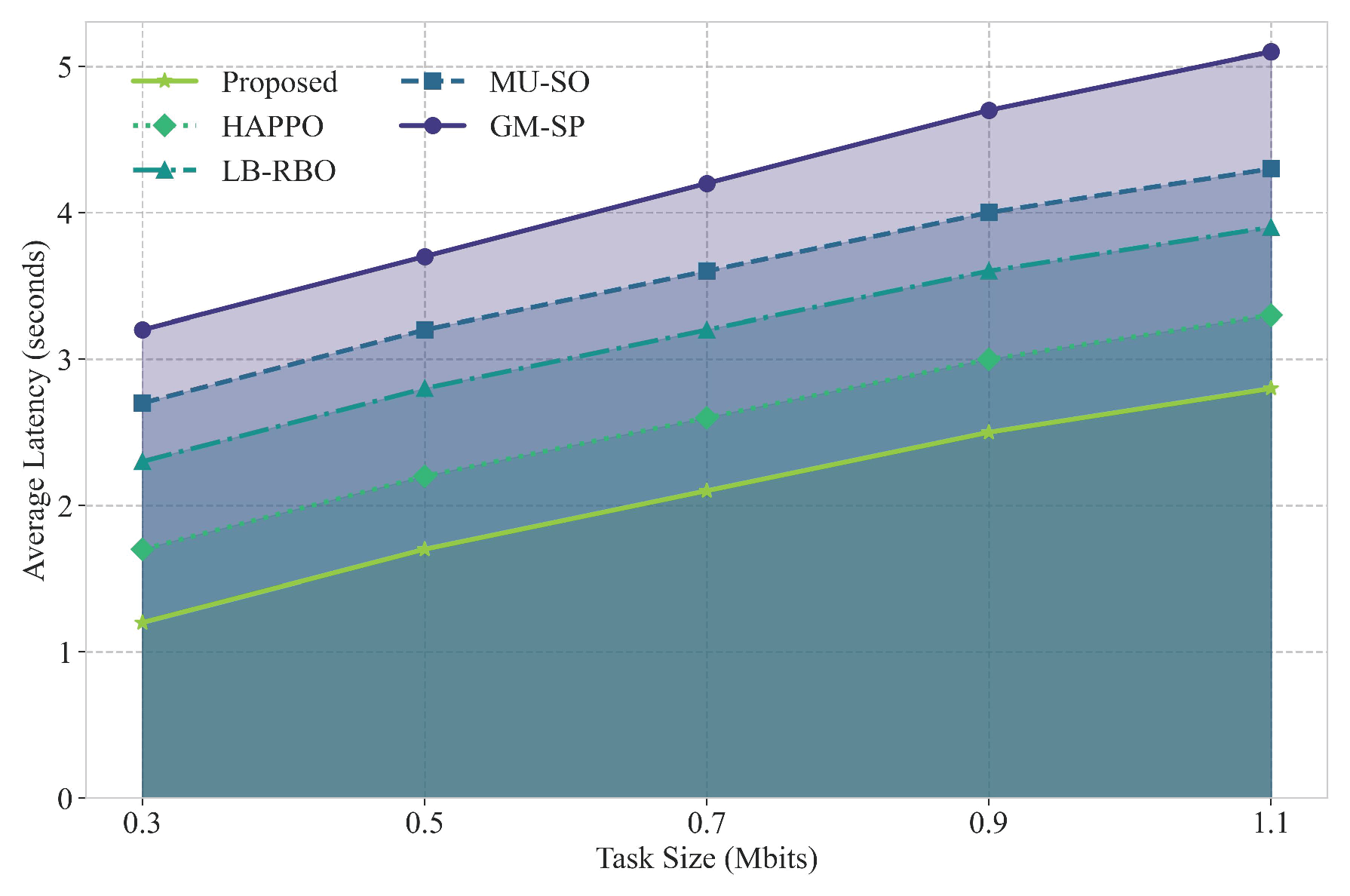}}
\caption{System performance under varying task sizes.}
\label{f6}
\end{figure*}

\subsection{Impact of UAV Density on System Performance}

To evaluate the scalability of the proposed framework, we conduct a series of experiments by varying the number of UAVs from 2 to 10. Fig.~\ref{f5} presents the system performance in terms of energy efficiency, task completion ratio, and average task latency, compared against four baseline algorithms. At 10 UAVs, the proposed framework achieves over 22,000~bit/J, representing approximately a 10\% improvement over the second-best method, with a task completion rate about 5\% higher and an average latency reduction of roughly 17\%.

As shown in Fig.~\ref{f5}(a), the proposed framework consistently achieves the highest energy efficiency across all UAV numbers. As the number of UAVs increases, the efficiency gains become more pronounced, reaching over 22,000 bit/J at 10 UAVs-significantly higher than the second-best baseline, HAPPO. This demonstrates the superiority of the proposed framework in leveraging additional UAVs for cooperative offloading and energy-aware decision making.

In Fig.~\ref{f5}(b), the proposed framework achieves the highest task completion ratio, exceeding 90\% when 10 UAVs are deployed. This is followed by HAPPO, which reaches about 85\%, while rule-based strategies (MU-SO, LB-RBO) lag behind. GM-SP performs the worst, showing only modest improvements with additional UAVs. The superior performance of the proposed framework reflects its ability to adaptively coordinate task allocation under dynamic network loads.

Fig.~\ref{f5}(c) shows that the proposed framework results in the lowest average task latency, decreasing from about 2.3s at 2 UAVs to nearly 1.5s at 10 UAVs. HAPPO also reduces the latency with more UAVs but remains consistently higher than the proposed framework. Static strategies such as GM-SP and MU-SO exhibit higher and flatter latency trends, indicating poor responsiveness to increased parallelism. These results validate the effectiveness of diffusion-enhanced decision-making in minimizing delay while scaling with UAV density.

\begin{figure*}[htbp]
\centering
\subfloat[Energy efficiency vs. communication bandwidth.]{
\includegraphics[width=5.8cm]{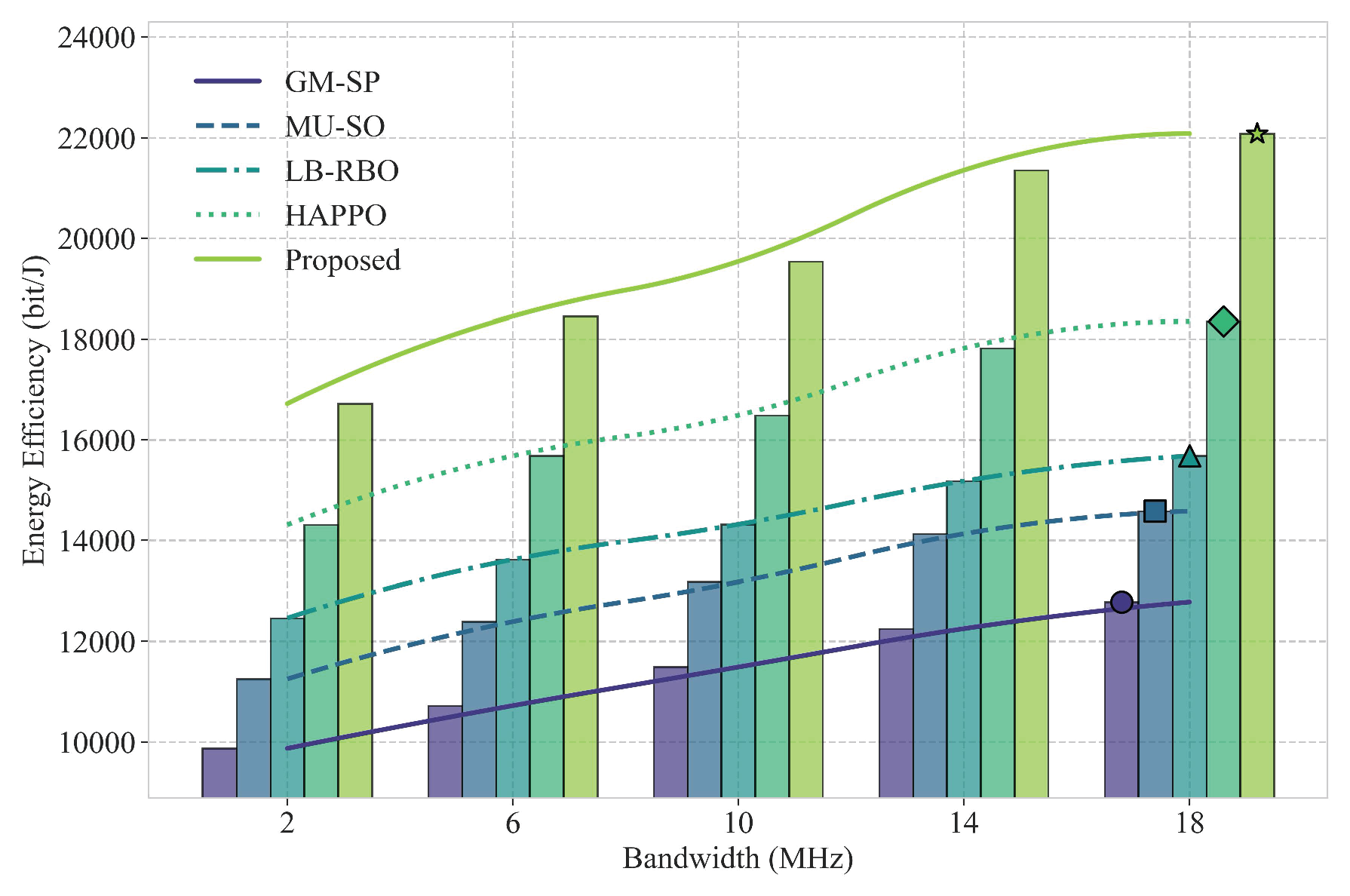}}
\hfill
\subfloat[Task completion ratio vs. communication bandwidth.]{
\includegraphics[width=5.8cm]{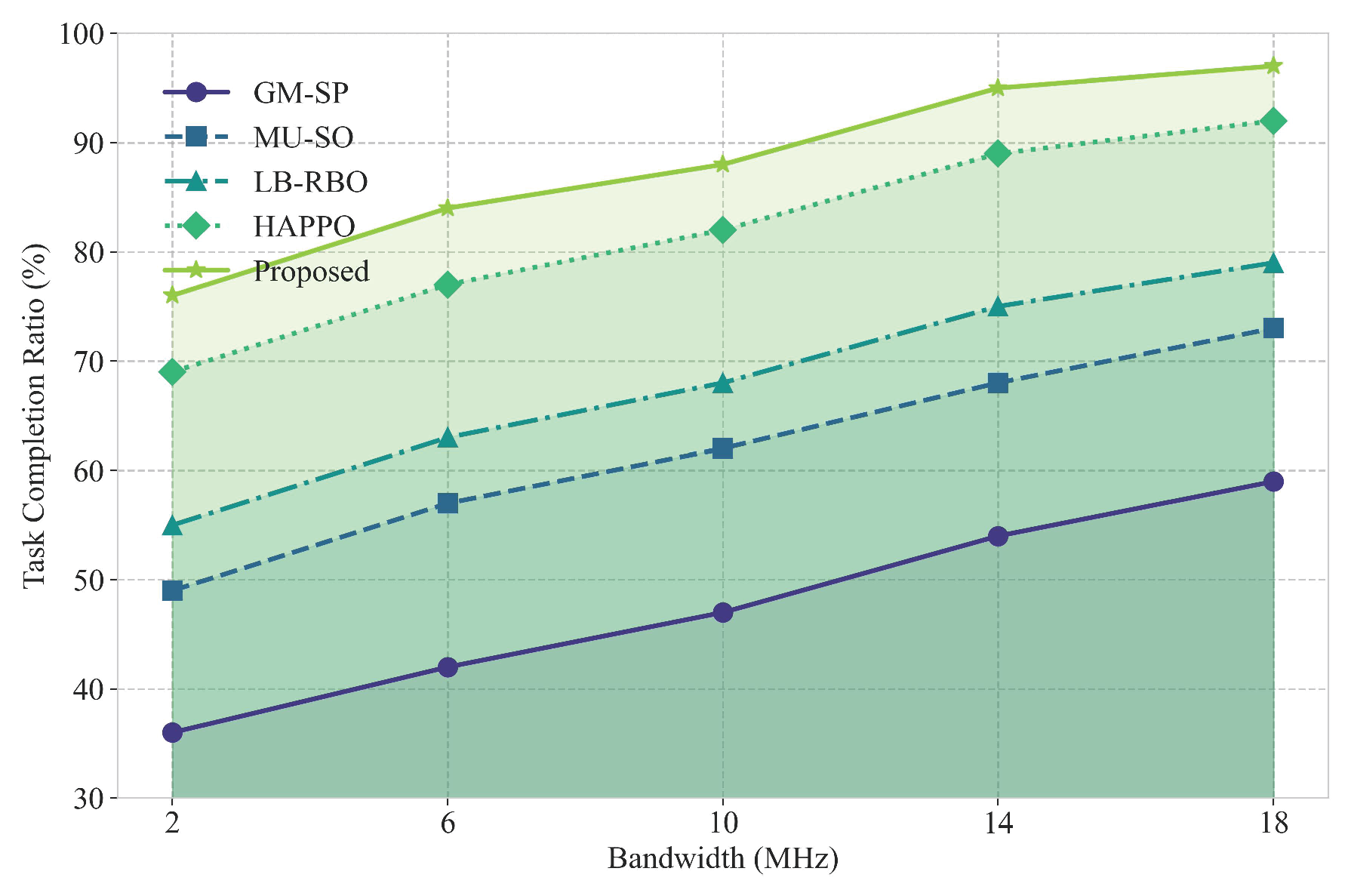}}
\hfill
\subfloat[Average task latency vs. communication bandwidth.]{
\includegraphics[width=5.8cm]{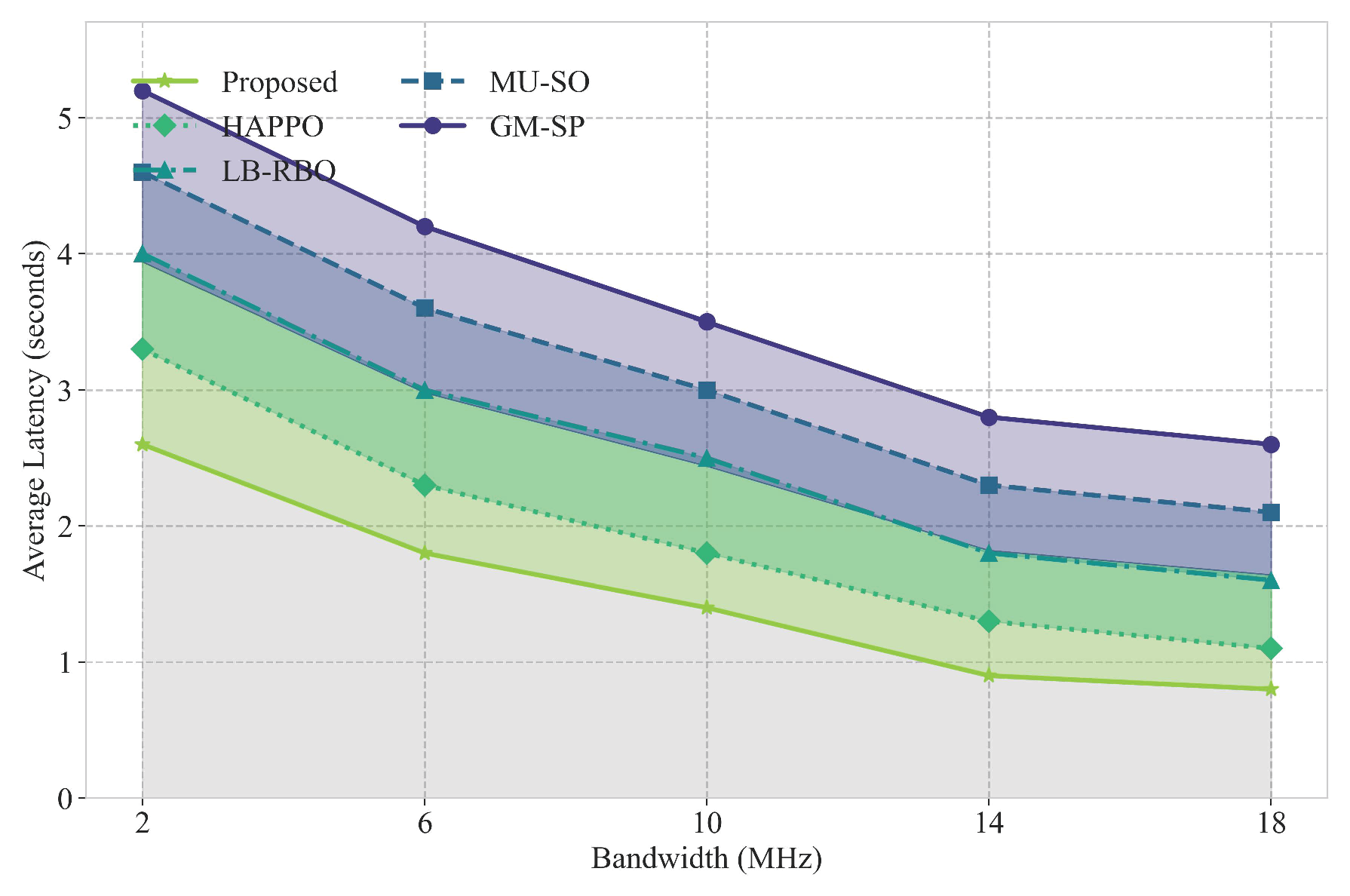}}
\caption{System performance under varying communication bandwidths.}
\label{f7}
\end{figure*}

\subsection{Impact of Task Size on System Performance}

To evaluate how task workload affects system performance, we vary the task size from 0.3 Mbits to 1.1 Mbits and present the results in Fig.~\ref{f6}. Even at the largest task size of 1.1~Mbits, the proposed framework maintains more than 75\% task completion, outperforming the best baseline by over 8\% and achieving an energy efficiency improvement of approximately 12\%.

As shown in Fig.~\ref{f6}(a), the energy efficiency of all algorithms decreases as task size increases, due to higher computation and transmission overheads. However, the proposed framework consistently achieves the highest efficiency across all task sizes, benefiting from its ability to dynamically select energy-optimal offloading strategies based on real-time conditions.

Fig.~\ref{f6}(b) illustrates that the task completion ratio drops as tasks become larger, reflecting the growing difficulty in meeting deadlines. The proposed framework significantly outperforms all baselines, maintaining over 75\% completion even at 1.1 Mbits. Its performance gap becomes more evident with increasing task size, demonstrating superior adaptability under heavy workloads.

In Fig.~\ref{f6}(c), the average task latency rises across all methods as the task size increases. The proposed framework maintains the lowest latency profile throughout, due to effective load balancing and queue-aware scheduling. In contrast, static policies such as GM-SP and MU-SO suffer from growing queuing delays and lack of adaptive routing, resulting in substantially higher latency under large tasks.

\subsection{Impact of Communication Bandwidth on System Performance}

To evaluate the influence of communication capacity on system performance, we vary the available bandwidth from 2 MHz to 18 MHz and summarize the results in Fig.~\ref{f7}. At 18~MHz, it achieves over 90\% task completion, 6\% higher than HAPPO, while sustaining an energy efficiency advantage of roughly 9\%.

As shown in Fig.~\ref{f7}(a), the energy efficiency improves consistently with increasing bandwidth for all methods, due to reduced transmission delays and energy overheads. The proposed framework exhibits the most significant gains, leveraging its dynamic offloading policy to fully utilize enhanced link capacities. In contrast, the rule-based baselines such as MU-SO and GM-SP improve slowly due to fixed offloading thresholds and lack of real-time channel adaptation.

In Fig.~\ref{f7}(b), the task completion ratio rises across all schemes with wider bandwidth. The proposed framework achieves the highest completion rates, surpassing 90\% at 18 MHz. It outperforms HAPPO and other baselines by a wide margin, particularly under constrained bandwidth conditions, demonstrating superior responsiveness to varying transmission capacity.

Fig.~\ref{f7}(c) illustrates that the average latency drops as the bandwidth increases, due to faster transmission and reduced queuing. The proposed framework maintains the lowest latency throughout the bandwidth range, remaining below 2 seconds even at low-bandwidth settings. This confirms the advantage of its adaptive scheduling mechanism in maintaining fast response time under limited wireless resources.

\section{CONCLUSIONS}\label{s6}
In this paper, we proposed a hierarchical task offloading and trajectory optimization framework for LAINs, addressing the challenges of dynamic task arrivals, energy-constrained UAVs, and heterogeneous edge resources. To manage the complexity of the joint optimization problem, we introduced a two-timescale solution architecture that integrates auction theory with generative MARL.
At the large timescale, a VCG auction mechanism was developed to perform energy-efficient UAV-task assignment, ensuring incentive-compatible and utility-maximizing allocation decisions. At the small timescale, we proposed D-HAPPO, a diffusion-enhanced heterogeneous-agent PPO algorithm. By embedding LDMs into the actor networks, D-HAPPO enabled conditional action generation under uncertainty, thereby improving the decision diversity, learning stability, and responsiveness in stochastic environments.
Extensive experiments under diverse simulation scenarios demonstrated that the proposed method consistently outperformed existing baselines in terms of energy efficiency, task completion ratio, and latency. Notably, D-HAPPO maintained high robustness across varying UAV densities, task complexities, and communication bandwidths, validating its scalability and adaptability in realistic deployment settings.

\bibliography{references}

\begin{thebibliography}{10}
\providecommand{\url}[1]{#1}
\csname url@samestyle\endcsname
\providecommand{\newblock}{\relax}
\providecommand{\bibinfo}[2]{#2}
\providecommand{\BIBentrySTDinterwordspacing}{\spaceskip=0pt\relax}
\providecommand{\BIBentryALTinterwordstretchfactor}{4}
\providecommand{\BIBentryALTinterwordspacing}{\spaceskip=\fontdimen2\font plus
\BIBentryALTinterwordstretchfactor\fontdimen3\font minus
  \fontdimen4\font\relax}
\providecommand{\BIBforeignlanguage}[2]{{%
\expandafter\ifx\csname l@#1\endcsname\relax
\typeout{** WARNING: IEEEtran.bst: No hyphenation pattern has been}%
\typeout{** loaded for the language `#1'. Using the pattern for}%
\typeout{** the default language instead.}%
\else
\language=\csname l@#1\endcsname
\fi
#2}}
\providecommand{\BIBdecl}{\relax}
\BIBdecl

\bibitem{EETO_2022}
L.~Zhang, A.~Celik, S.~Dang, and B.~Shihada, ``Energy-efficient trajectory
  optimization for {UAV}-assisted {IoT} networks,'' \emph{IEEE Trans. Mob.
  Comput.}, vol.~21, no.~12, pp. 4323--4337, Dec. 2022.

\bibitem{3DTO_2022}
B.~Li, Q.~Li, Y.~Zeng, Y.~Rong, and R.~Zhang, ``{3D} trajectory optimization
  for energy-efficient {UAV} communication: A control design perspective,''
  \emph{IEEE Trans. Wireless Commun.}, vol.~21, no.~6, pp. 4579--4593, Jun.
  2022.

\bibitem{SFCD_2025}
Z.~Jia, Y.~Cao, L.~He, Q.~Wu, Q.~Zhu, D.~Niyato, and Z.~Han, ``Service function
  chain dynamic scheduling in space-air-ground integrated networks,''
  \emph{IEEE Trans. Veh. Technol.}, vol.~74, no.~7, pp. 11\,235--11\,248, Jul.
  2025.

\bibitem{UTOT_2022}
K.~Liu and J.~Zheng, ``{UAV} trajectory optimization for time-constrained data
  collection in {UAV}-enabled environmental monitoring systems,'' \emph{IEEE
  Internet Things J.}, vol.~23, no.~9, pp. 24\,300--24\,314, Dec. 2022.

\bibitem{AETO_2024}
P.~Du, Y.~Shi, H.~Cao, S.~Garg, M.~Alrashoud, and P.~K. Shukla, ``{AI}-enabled
  trajectory optimization of logistics {UAVs} with wind impacts in smart
  cities,'' \emph{IEEE Trans. Consum. Electron.}, vol.~70, no.~1, pp.
  3885--3897, Feb. 2024.

\bibitem{Sustainable_Wang_2024}
F.~Wang, S.~Zhang, J.~Shi, Z.~Li, and T.~Q.~S. Quek, ``Sustainable {UAV}
  mobility support in integrated terrestrial and non-terrestrial networks,''
  \emph{IEEE Trans. Wireless Commun.}, vol.~23, no.~11, pp. 17\,115--17\,128,
  Nov. 2024.

\bibitem{NFV_cyl_2025}
Z.~Jia, Y.~Cao, L.~He, G.~Li, F.~Zhou, Q.~Wu, and Z.~Han, ``{NFV}-enabled
  service recovery in space-air-ground integrated networks: A matching
  game-based approach,'' \emph{IEEE Trans. Netw. Sci. Eng.}, vol.~12, no.~3,
  pp. 1732--1744, May 2025.

\bibitem{Joint_zhao_2025}
Y.~Zhao, C.~Liu, X.~Hu, J.~He, M.~Peng, D.~Wing Kwan~Ng, and T.~Q.~S. Quek,
  ``Joint content caching, service placement, and task offloading in
  {UAV}-enabled mobile edge computing networks,'' \emph{IEEE J. Sel. Areas
  Commun.}, vol.~43, no.~1, pp. 51--63, Jan. 2025.

\bibitem{Chen_2024}
W.~Chen, C.~Liu, W.~Wang, M.~Peng, and W.~Zhang, ``Adaptive hybrid beamforming
  for uav mmwave communications against asymmetric jitter,'' \emph{IEEE Trans.
  Wireless Commun.}, vol.~23, no.~8, pp. 9432--9445, Aug. 2024.

\bibitem{EATD_2022}
N.~Gupta, D.~Mishra, and S.~Agarwal, ``Energy-aware trajectory design for
  outage minimization in {UAV}-assisted communication systems,'' \emph{IEEE
  Trans. Green Commun. Netw.}, vol.~6, no.~3, pp. 1751--1763, Sep. 2022.

\bibitem{G2S_2025_arxiv}
W.~Yuan, Y.~Cui, J.~Wang, F.~Liu, G.~Sun, T.~Xiang, J.~Xu, S.~Jin, D.~Niyato,
  S.~Coleri, S.~Sun, S.~Mao, A.~Jamalipour, D.~I. Kim, M.-S. Alouini, and
  X.~Shen, ``From ground to sky: Architectures, applications, and challenges
  shaping low-altitude wireless networks,'' \emph{arXiv e-prints
  arXiv:2506.12308}, 2025.

\bibitem{Disrupting_2025_arxiv}
C.~Xie, J.~He, S.~Guo, J.~Wang, S.~Zhang, T.~Zhang, and T.~Xiang, ``Disrupting
  vision-language model-driven navigation services via adversarial object
  fusion,'' \emph{arXiv e-prints arXiv:2505.23266}, 2025.

\bibitem{CWDU_2023}
Z.~Wang and L.~Duan, ``Chase or wait: Dynamic {UAV} deployment to learn and
  catch time-varying user activities,'' \emph{IEEE Trans. Mob. Comput.},
  vol.~22, no.~3, pp. 1369--1383, Mar. 2023.

\bibitem{TVBM_2024}
H.~Wei and H.~Zhang, ``Time-varying boundary modeling and handover analysis of
  {UAV}-assisted networks with fading,'' \emph{IEEE Trans. Wireless Commun.},
  vol.~23, no.~7, pp. 1369--1383, Jul. 2024.

\bibitem{DROA_2025}
Z.~Jia, C.~Cui, C.~Dong, Q.~Wu, Z.~Ling, D.~Niyato, and Z.~Han,
  ``Distributionally robust optimization for aerial multi-access edge computing
  via cooperation of {UAV}s and {HAP}s,'' \emph{IEEE Trans. Mob. Comput.},
  vol.~24, no.~10, pp. 10\,853--10\,867, Oct 2025.

\bibitem{Cooperative_2024}
Z.~Jia, J.~You, C.~Dong, Q.~Wu, F.~Zhou, D.~Niyato, and Z.~Han, ``Cooperative
  cognitive dynamic system in {UAV} swarms: Reconfigurable mechanism and
  framework,'' \emph{IEEE Veh. Technol. Mag.}, vol.~19, no.~3, pp. 90--101,
  Jul. 2024.

\bibitem{VCG_2005}
A.~Karlin and D.~Kempe, ``Beyond {VCG}: frugality of truthful mechanisms,'' in
  \emph{46th Annual IEEE Symposium on Foundations of Computer Science
  (FOCS'05)}, Pittsburgh, PA, USA, Oct. 2005.

\bibitem{VCG_2023}
L.~Wu, S.~Guo, Z.~Hong, Y.~Liu, W.~Xu, and Y.~Zhan, ``Long-term adaptive {VCG}
  auction mechanism for sustainable federated learning with periodical client
  shifting,'' \emph{IEEE Trans. Mob. Comput.}, vol.~23, no.~5, pp. 6060--6073,
  May 2023.

\bibitem{Two_based_2024}
F.~Zhao, B.~Yang, C.~Li, C.~Zhang, L.~Zhu, and G.~Liang, ``Two-layer consensus
  based on primary-secondary consortium chain data sharing for internet of
  vehicles,'' \emph{IEEE Trans. Veh. Technol.}, vol.~73, no.~9, pp.
  13\,828--13\,838, Sep. 2024.

\bibitem{LDM_2022}
R.~Rombach, A.~Blattmann, D.~Lorenz, P.~Esser, and B.~Ommer, ``High-resolution
  image synthesis with latent diffusion models,'' in \emph{Proceedings of the
  IEEE/CVF Conference on Computer Vision and Pattern Recognition (CVPR)}, New
  Orleans, Louisiana, Aug. 2022.

\bibitem{Aerial_2025_sun}
G.~Sun, J.~Xiao, J.~Li, J.~Wang, J.~Kang, D.~Niyato, and S.~Mao, ``Aerial
  reliable collaborative communications for terrestrial mobile users via
  evolutionary multi-objective deep reinforcement learning,'' \emph{IEEE Trans.
  Mob. Comput.}, vol.~24, no.~7, pp. 5731--5748, Jul. 2025.

\bibitem{Generative_2025_Wang}
J.~Wang, H.~Du, D.~Niyato, J.~Kang, S.~Cui, X.~Shen, and P.~Zhang, ``Generative
  {AI} for integrated sensing and communication: Insights from the physical
  layer perspective,'' \emph{IEEE Wirel. Commun.}, vol.~31, no.~5, pp.
  246--255, Oct. 2024.

\bibitem{Computational_Cao_2024}
P.~Cao, L.~Lei, S.~Cai, G.~Shen, X.~Liu, X.~Wang, L.~Zhang, L.~Zhou, and
  M.~Guizani, ``Computational intelligence algorithms for {UAV} swarm
  networking and collaboration: A comprehensive survey and future directions,''
  \emph{IEEE Commun. Surv. Tutor.}, vol.~26, no.~4, pp. 2684--2728, 4th Quart.
  2024.

\bibitem{Revolutionizing_Mahboob_2024}
S.~Mahboob and L.~Liu, ``Revolutionizing future connectivity: A contemporary
  survey on {AI}-empowered satellite-based non-terrestrial networks in {6G},''
  \emph{IEEE Commun. Surv. Tutor.}, vol.~26, no.~2, pp. 1279--1321, 2nd Quart.
  2024.

\bibitem{Cooperative_2025_low}
J.~Tang, Y.~Yu, C.~Pan, H.~Ren, D.~Wang, J.~Wang, and X.~You, ``Cooperative
  {ISAC}-empowered low-altitude economy,'' \emph{IEEE Trans. Wireless Commun.},
  Feb. 2025, early access.

\bibitem{Airspace_2024}
H.~Huang, J.~Su, and F.-Y. Wang, ``The potential of low-altitude airspace: The
  future of urban air transportation,'' \emph{IEEE Trans. Intell. Veh.},
  vol.~9, no.~8, pp. 5250--5254, Aug. 2024.

\bibitem{Sensitive_2024}
X.~Gao, X.~Zhu, and L.~Zhai, ``{AoI}-sensitive data collection in
  multi-{UAV}-assisted wireless sensor networks,'' \emph{IEEE Trans. Wireless
  Commun.}, vol.~22, no.~8, pp. 5185--5197, Aug. 2023.

\bibitem{Multi_2024}
H.~Guo, Y.~Wang, J.~Liu, and C.~Liu, ``Multi-{UAV} cooperative task offloading
  and resource allocation in {5G} advanced and beyond,'' \emph{IEEE Trans.
  Wireless Commun.}, vol.~23, no.~1, pp. 347--359, Jan. 2024.

\bibitem{Secure_2023}
Y.~Wang, Z.~Su, Q.~Xu, R.~Li, T.~H. Luan, and P.~Wang, ``A secure and
  intelligent data sharing scheme for {UAV}-assisted disaster rescue,''
  \emph{IEEE/ACM Trans. Netw.}, vol.~31, no.~6, pp. 2422--2438, Dec. 2023.

\bibitem{Navigation_2024}
D.~Dissanayaka, T.~R. Wanasinghe, O.~De~Silva, A.~Jayasiri, and G.~K.~I. Mann,
  ``Review of navigation methods for {UAV}-based parcel delivery,'' \emph{IEEE
  Trans. Autom. Sci. Eng.}, vol.~21, no.~1, pp. 1068--1082, Jan. 2024.

\bibitem{MultiAgent_2024}
Z.~Ning, Y.~Yang, X.~Wang, Q.~Song, L.~Guo, and A.~Jamalipour, ``Multi-agent
  deep reinforcement learning based {UAV} trajectory optimization for
  differentiated services,'' \emph{IEEE Trans. Mob. Comput.}, vol.~23, no.~5,
  pp. 5818--5834, May 2024.

\bibitem{VEC_2024}
X.~Dai, Z.~Xiao, H.~Jiang, and J.~C.~S. Lui, ``{UAV}-assisted task offloading
  in vehicular edge computing networks,'' \emph{IEEE Trans. Mob. Comput.},
  vol.~23, no.~4, pp. 2520--2534, Apr. 2024.

\bibitem{ECE_2024}
Y.~Qu, H.~Sun, C.~Dong, J.~Kang, H.~Dai, Q.~Wu, and S.~Guo, ``Elastic
  collaborative edge intelligence for {UAV} swarm: Architecture, challenges,
  and opportunities,'' \emph{IEEE Commun. Mag.}, vol.~62, no.~1, pp. 62--68,
  Jan. 2024.

\bibitem{TOEP_2024}
A.~Gao, S.~Zhang, Q.~Zhang, Y.~Hu, S.~Liu, W.~Liang, and S.~X. Ng, ``Task
  offloading and energy optimization in hybrid {UAV}-assisted mobile edge
  computing systems,'' \emph{IEEE Trans. Veh. Technol.}, vol.~73, no.~8, pp.
  12\,052--12\,066, Aug. 2024.

\bibitem{ELEJ_2024}
F.~Pervez, A.~Sultana, C.~Yang, and L.~Zhao, ``Energy and latency efficient
  joint communication and computation optimization in a multi-{UAV}-assisted
  {MEC} network,'' \emph{IEEE Trans. Wireless Commun.}, vol.~23, no.~3, pp.
  1728--1741, Mar. 2024.

\bibitem{APAF_2023}
N.~Lin, H.~Tang, L.~Zhao, S.~Wan, A.~Hawbani, and M.~Guizani, ``A {PDDQNLP}
  algorithm for energy efficient computation offloading in {UAV}-assisted
  {MEC},'' \emph{IEEE Trans. Wireless Commun.}, vol.~22, no.~12, pp.
  8876--8890, Dec. 2023.

\bibitem{EETO_2024}
F.~Song, M.~Deng, H.~Xing, Y.~Liu, F.~Ye, and Z.~Xiao, ``Energy-efficient
  trajectory optimization with wireless charging in {UAV}-assisted {MEC} based
  on multi-objective reinforcement learning,'' \emph{IEEE Trans. Mob. Comput.},
  vol.~23, no.~12, pp. 10\,867--10\,884, Dec. 2024.

\bibitem{MODO_2024}
X.~Zhu, L.~Zhai, N.~Li, Y.~Li, and F.~Yang, ``Multi-objective deployment
  optimization of {UAVs} for energy-efficient wireless coverage,'' \emph{IEEE
  Trans. Commun.}, vol.~23, no.~12, pp. 10\,867--10\,884, Dec. 2024.

\bibitem{JECT_2024}
C.~Peng, Z.~Wu, X.~Huang, Y.~Wu, J.~Kang, Q.~Huang, and S.~Xie, ``Joint energy
  and completion time difference minimization for {UAV}-enabled intelligent
  transportation systems: A constrained multi-objective optimization
  approach,'' \emph{IEEE Trans. Intell. Transp. Syst.}, vol.~25, no.~10, pp.
  14\,040--14\,053, Oct. 2024.

\bibitem{RISA_2024}
L.~Zhai, Y.~Zou, J.~Zhu, and Y.~Jiang, ``{RIS}-assisted {UAV}-enabled wireless
  powered communications: System modeling and optimization,'' \emph{IEEE Trans.
  Wireless Commun.}, vol.~23, no.~5, pp. 5094--5108, May 2024.

\bibitem{TSSF_2024}
J.~Shi, P.~Cong, L.~Zhao, X.~Wang, S.~Wan, and M.~Guizani, ``A two-stage
  strategy for {UAV}-enabled wireless power transfer in unknown environments,''
  \emph{IEEE Trans. Mob. Comput.}, vol.~23, no.~2, pp. 1785--1802, Feb. 2024.

\bibitem{CFBG_2023}
N.~Qi, Z.~Huang, W.~Sun, S.~Jin, and X.~Su, ``Coalitional formation-based
  group-buying for {UAV}-enabled data collection: An auction game approach,''
  \emph{IEEE Trans. Mob. Comput.}, vol.~22, no.~12, pp. 7420--7437, Dec. 2023.

\bibitem{OAAT_2024}
K.~Mo, X.~Li, C.~J. Xue, Z.~Li, and H.~Xu, ``An online auction approach to
  {UAV} scheduling and trajectory planning,'' in \emph{ICC 2024 - IEEE
  International Conference on Communications}, Denver, CO, USA, Aug. 2024.

\bibitem{Diffusion_du_2024}
H.~Du, R.~Zhang, Y.~Liu, J.~Wang, Y.~Lin, Z.~Li, D.~Niyato, J.~Kang, Z.~Xiong,
  S.~Cui, B.~Ai, H.~Zhou, and D.~I. Kim, ``Enhancing deep reinforcement
  learning: A tutorial on generative diffusion models in network
  optimization,'' \emph{IEEE Commun. Surv. Tutor.}, vol.~26, no.~4, pp.
  2611--2646, 4th Quart. 2024.

\bibitem{IMOT_2024}
R.~Zhang, R.~Zhou, Y.~Wang, H.~Tan, and K.~He, ``Incentive mechanisms for
  online task offloading with privacy-preserving in {UAV}-assisted mobile edge
  computing,'' \emph{IEEE/ACM Trans. Netw.}, vol.~32, no.~3, pp. 2646--2661,
  Jun. 2024.

\bibitem{IOW_2023}
M.~Dai, Z.~Luo, Y.~Wu, L.~Qian, B.~Lin, and Z.~Su, ``Incentive oriented
  two-tier task offloading scheme in marine edge computing networks: A hybrid
  stackelberg-auction game approach,'' \emph{IEEE Trans. Wireless Commun.},
  vol.~22, no.~3, pp. 8603--8619, Dec. 2023.

\bibitem{DNN_2025}
Z.~Liu, H.~Du, J.~Lin, Z.~Gao, L.~Huang, S.~Hosseinalipour, and D.~Niyato,
  ``{DNN} partitioning, task offloading, and resource allocation in dynamic
  vehicular networks: A lyapunov-guided diffusion-based reinforcement learning
  approach,'' \emph{IEEE Trans. Mob. Comput.}, vol.~24, no.~3, pp. 1945--1962,
  Mar. 2025.

\bibitem{MOAC_2025}
C.~Zhang, G.~Sun, J.~Li, Q.~Wu, J.~Wang, D.~Niyato, and Y.~Liu,
  ``Multi-objective aerial collaborative secure communication optimization via
  generative diffusion model-enabled deep reinforcement learning,'' \emph{IEEE
  Trans. Mob. Comput.}, vol.~24, no.~4, pp. 3041--3058, Apr. 2025.

\bibitem{Joint_Optimization_Zhao_2025}
M.~Zhao, R.~Zhang, Z.~He, and K.~Li, ``Joint optimization of trajectory,
  offloading, caching, and migration for {UAV}-assisted {MEC},'' \emph{IEEE
  Trans. Mob. Comput.}, vol.~24, no.~12, pp. 1981--1998, Mar. 2025.

\bibitem{Tradeoff_Between_Zhan_2024}
C.~Zhan, H.~Hu, J.~Wang, Z.~Liu, and S.~Mao, ``Tradeoff between age of
  information and operation time for {UAV} sensing over multi-cell cellular
  networks,'' \emph{IEEE Trans. Mob. Comput.}, vol.~23, no.~4, pp. 2976--2991,
  Apr. 2024.

\bibitem{UAV_Enabled_Xu_2018}
J.~Xu, Y.~Zeng, and R.~Zhang, ``{UAV}-enabled wireless power transfer:
  Trajectory design and energy optimization,'' \emph{IEEE Trans. Wireless
  Commun.}, vol.~17, no.~8, pp. 5092--5106, Aug. 2018.

\bibitem{al_Ahmed_2023}
A.~A. Al-Habob, O.~A. Dobre, S.~Muhaidat, and H.~V. Poor, ``Energy-efficient
  information placement and delivery using {UAVs},'' \emph{IEEE Internet Things
  J.}, vol.~10, no.~1, pp. 357--366, Jan. 2023.

\bibitem{Energy_Mu_2021}
X.~Mu, Y.~Liu, L.~Guo, J.~Lin, and Z.~Ding, ``Energy-constrained {UAV} data
  collection systems: {NOMA} and {OMA},'' \emph{IEEE Trans. Veh. Technol.},
  vol.~70, no.~7, pp. 6898--6912, Jul. 2021.

\bibitem{UEE_2024}
J.~Su, Z.~Liu, Y.-a. Xie, Y.~Li, K.~Ma, and X.~Guan, ``{UEE}–delay balanced
  online resource optimization for cooperative {MEC}-enabled task offloading in
  dynamic vehicular networks,'' \emph{IEEE Internet Things J.}, vol.~11, no.~7,
  pp. 11\,496--11\,507, Apr. 2023.

\bibitem{MLTE_2025}
Z.~Yao, Q.~Zhu, Y.~Zhang, H.~Huang, and M.~Luo, ``Minimizing long-term energy
  consumption in {RIS}-assisted {UAV}-enabled {MEC} network,'' \emph{IEEE
  Internet Things J.}, Feb. 2025, early access.

\bibitem{HARL_2024}
Y.~Zhong, J.~G. Kuba, X.~Feng, S.~Hu, J.~Ji, and Y.~Yang, ``Heterogeneous-agent
  reinforcement learning,'' \emph{J. Mach. Learn. Res.}, vol.~25, no.~32, pp.
  1--67, Jan. 2024.

\bibitem{JCOR_2025}
J.~You, Z.~Jia, C.~Dong, Q.~Wu, and Z.~Han, ``Joint computation offloading and
  resource allocation for uncertain maritime {MEC} via cooperation of {UAVs}
  and vessels,'' \emph{IEEE Trans. Veh. Technol.}, Jun. 2025, early access.

\end{thebibliography}

\end{document}